\pdfoutput=1
\documentclass[review]{elsarticle}
\usepackage{subfigure}
\usepackage{longtable}
\usepackage{algorithm, algorithmic}
\usepackage{bm}
\usepackage{booktabs}
\usepackage{multirow}
\usepackage{geometry}
\usepackage{amsfonts,amssymb}

\usepackage{lineno,hyperref}
\modulolinenumbers[5]

\journal{arXiv}

\begin{document}

\begin{frontmatter}

\title{Multi-granularity Relabeled Under-sampling Algorithm for Imbalanced Data}


\author[mymainaddress]{Qi Dai}

\author[mymainaddress]{Jian-wei Liu\corref{mycorrespondingauthor}}
\cortext[mycorrespondingauthor]{Corresponding author}
\ead{liujw@cup.edu.cn}

\author[mainaddress]{Yang Liu}

\address[mymainaddress]{Department of Automation, College of Information Science and Engineering,
China University of Petroleum , Beijing, Beijing, China}

\address[mainaddress]{College of Science, North China University of Science and Technology (NCST), Tangshan, China}

\begin{abstract}
The imbalanced classification problem turns out to be one of the important and challenging problems in data mining and machine learning. The performances of traditional classifiers will be severely affected by many data problems, such as class imbalanced problem, class overlap and noise. When the number of one class in the data set is larger than other classes, class imbalanced problem will inevitably occur. Therefore, many researchers are committed to solving the problem of category imbalance and improving the overall classification performances of the classifier. The Tomek-Link algorithm was only used to clean data when it was proposed. In recent years, there have been reports of combining Tomek-Link algorithm with sampling technique. The Tomek-Link sampling algorithm can effectively reduce the class overlap on data, remove the majority instances that are difficult to distinguish, and improve the algorithm classification accuracy. However, the Tomek-Links under-sampling algorithm only considers the boundary instances that are the nearest neighbors to each other globally and ignores the potential local overlapping instances. When the number of minority instances is small, the under-sampling effect is not satisfactory, and the performance improvement of the classification model is not obvious. Therefore, on the basis of Tomek-Link, a multi-granularity relabeled under-sampling algorithm (MGRU) is proposed. This algorithm fully considers the local information of the data set in the local granularity subspace, and detects the local potential overlapping instances in the data set. Then, the overlapped majority instances are eliminated according to the global relabeled index value, which effectively expands the detection range of Tomek-Links. The simulation results show that when we select the optimal global relabeled index value for under-sampling, the classification accuracy and generalization performance of the proposed under-sampling algorithm are significantly better than other baseline algorithms.
\end{abstract}


\end{frontmatter}

\section{Introduction}

Imbalanced data is a special form of existence in the data mining field. It has high research value or commercial significance in many application fields, for example, network intrusion detection \cite{1khor2012cascaded,2garcia2012oligois,3bamakan2017ramp}, fault detection \cite{4kwak2015incremental}, software defect prediction \cite{5sun2020collaborative}, spam review detection \cite{6jin2015filtering} and other fields. In a binary-class imbalanced data set, the performances of traditional classifiers are easily affected seriously when the number of instances in the negative (majority) class is overwhelmed the number of instances in the positive (minority) class \cite{7sun2009classification}. Traditional classifiers usually assume that the distribution of class in the data set is balanced. In the process of imbalanced data classification, the overall classification accuracy may be high, while classification accuracy of the positive classes is low \cite{8haixiang2017learning}. Imbalanced data has the two structural characteristics, i.e., imbalanced distribution and class overlap. The traditional classification model cannot effectively handle the structural characteristics existing in imbalanced data. It is difficult to determine the true classification boundary of the data set, resulting in low recognition accuracy of positive instances \cite{9sun2007cost}.

In real world applications, positive instances often have higher value, while traditional classifiers cannot effectively identify positive instances in imbalanced data, and lose the significance of classification learning on imbalanced data sets \cite{10he2009learning,11chen2018synthetic}. Till now imbalanced data mining algorithms roughly contain three types of studying approaches: data-level \cite{12shen2021new,13szlobodnyik2021data,14borowska2019rough}, algorithm-level \cite{15zhu2019multiple,16raghuwanshi2018class} and hybrid mining algorithms \cite{17wang2020imbalanced}. The data-level algorithms take different measures to change the data set distribution under classes to achieve data rebalance, thereby improving the classification performance of the model. The dominant data-level algorithms comprise three sorts: under-sampling, over-sampling and hybrid sampling \cite{18ng2014diversified,19wang2014resampling,20gazzah2015hybrid}. The algorithm level algorithm is mainly to modify the existing classification algorithm, so that the classifier can adapt to the data structure of imbalanced data, enhance the ability of the classifier to recognize positive instances, and improve the classification accuracy and generalization ability of the model, such as ensemble learning algorithm or Fuzzy weighted support vector machine, etc\cite{21jian2016new}. The hybrid mining algorithm uses both the data-level and the algorithm-level algorithms at the same time to maximize the advantages of the two algorithms.

In imbalanced classification problems, when multiple types of instances share a specific area in the data space, the phenomenon of class overlap will occur \cite{22das2018handling}. Since the different degree of overlap of the data sets, the number of overlapping instances in the data sets also has significant differences. And what makes things worse is that, although these instances belong to different categories, values of their feature are similar, and this complexity is a major obstacle to be overcome for imbalanced classification tasks \cite{23garcia2008k}. What further increases the difficulty is that, overlapping instances are often located near the decision boundary, and are likely to be incorrectly classified by the algorithm. In sum up, the existence of overlapping instances is not conducive to the classification effectiveness of the classifier and degrades the classification performance of the classifier.

In this paper, we will focus on sampling algorithms because they are simple and more flexible. In 1976, Tomek Ivan proposed a hybrid form of CNN \cite{24tomek1976two} and Tomek-Link algorithms to delete data from the data set. In recent years, many researchers have used the Tomek-Link algorithm for imbalanced data sampling, but the proposed algorithm is often combined with other algorithms \cite{25devi2017redundancy}. As far as we know, few researchers use the Tomek-Link algorithm as a single sampling algorithm to balance the data set. If the Tomek-Link algorithm is used as the sampling algorithm, this algorithm only considers the boundary instances that are the nearest neighbors to each other globally and ignores the potential local overlapping instances. Most of the class-overlap under-sampling algorithms proposed at present are suggested to search for global overlap instances in the entire data set through the nearest neighbor-based method. These algorithms must entail that the probability of feature overlap in the data set is the same. However, for real-world data sets, the probability of feature overlap is rarely the same. Therefore, if we simply consider the global overlapping instances, it is easy to ignore the potential overlapping instances in the data set, it may turn out that, such classification models are subjects to limited upgrading performance. To the best of our knowledge, our proposed algorithm is the first to consider locally overlapping instances in a local granular subspace. 

The main contributions of this paper are as follows:

(1) The proposed MGRU overcomes the problem that the UCBSS algorithm presented in \cite{53xing2018study} cannot be accommodated for real data sets.

(2) We expand the detection range of the Tomek-Links algorithm for overlapping instances.

(3) MGRU is the first algorithm that considers potentially overlapping majority instances in a local granularity subspace.

(4) We found that Mahalanobis distance and standardized Euclidean distance have little difference in performance when discovering potential-overlap instances in a local granularity subspace, and they can be used interchangeably.

The remainder of this manuscript is organized as follows. In Section 2, the relate work are briefly reviewed, the framework of the algorithm is introduced in section 3, the proposed algorithm is presented in Section 4, the experimental results and analyses are given in Section 5. Finally, Section 6 concludes this paper.

\section{Related work}

This research mainly discusses sampling algorithms of imbalanced data. Therefore, the research progress about sampling algorithms is mainly introduced in this section. The under-sampling algorithm improves the performance of the classification algorithm by deleting redundant or overlapping negative instances. Note that if negative instances are deleted indifferently, it may cause serious information loss on negative instances \cite{26koziarski2020combined}. The oversampling algorithm enhances the ability of the algorithm to identify positive instances by expanding the characteristics of positive instances. Although the oversampling algorithm alleviates the influence of imbalanced data on the algorithm to a certain extent, the oversampling algorithm may suffer from overfitting \cite{27oh2019oversampling}. The hybrid sampling algorithm combines the advantages of under-sampling and over-sampling algorithms to try to overcome the impact of the disadvantages of sampling algorithm itself on the classification performance of the algorithm \cite{28santos2018cross}. Next, we will focus on the classic under-sampling and class overlap algorithms.
With the increasing maturity of sampling technology, random algorithms such as random under-sampling are gradually replaced by other sampling algorithms due to their strong randomness and irrational removal of instances \cite{29cohen2006learning}. The under-sampling technique is mainly to use the algorithm to reasonably delete the redundant instances or overlap instances in the negative instance set, so that the model classification surface gradually shifts to the positive class. However, the under-sampling algorithm may suffer from the risk of unreasonably removing instances and loss of important information. 
Up to now, a large number of under-sampling algorithms have been proposed. We can roughly divide it into two categories: k-nearest neighbor-based algorithms and cluster-based algorithms \cite{30xie2021novel}. Most under-sampling algorithms remove majority instances based on k-nearest neighbors. Its purpose is to remove redundant instances or overlapping instances. Kubat et al. \cite{31kubat1997addressing} propose a one-sided selection (OSS) algorithm, which is an improved algorithm based on Tomek-Links. Hart et al. \cite{32hart1968condensed} combine Tomek-Links and propose a condensed nearest neighbor (CNN) algorithm, which treats majority instances in Tomek-Links pairs as either borderline or noise. Meanwhile, they delete the majority instances that are 1NN from each other in the majority instances, because they think such instances are redundant. Different from the above algorithm, Laurikkala et al. \cite{33laurikkala2001improving} propose the neighborhood cleaning rule (NCL) to remove majority instances based on the edited nearest neighbor (ENN) algorithm \cite{34tomek1976experiment}. And recently, related under-sampling algorithms have also been proposed. Devi et al. \cite{35devi2017redundancy} combine nearest neighbor and Tomek-Links under-sampling techniques, an improved under-sampling algorithm (TLUS) is proposed to be employed in the pre-processing stage. Kumar et al. \cite{36kumar2019tlusboost} propose an under-sampling ensemble classification algorithm (TLUSBoost) based on TLUS, which further improved the classification performances of TLUS under-sampling algorithm. Vuttipittayamongkol et al. \cite{37vuttipittayamongkol2020neighbourhood} propose an under-sampling framework to eliminate overlapping instances, which maximizes the visibility of minority instances and reduces the excessive elimination of data. They also propose four under-sampling algorithms under this framework.
Clustering algorithms are widely used by under-sampling algorithms. Yen et al. \cite{38yen2009cluster} propose an under-sampling algorithm based on k-means clustering, which randomly selects a sufficient number of instances from each cluster to effectively balance the training set. Lin et al. \cite{39lin2017clustering} propose a new under-sampling algorithm (Cluster-NN) based on the clustering algorithm, combined with 1NN technology to find the nearest neighbor instances of the cluster center. Ofek et al. \cite{40ofek2017fast} propose a fast clustering under-sampling algorithm (Fast-CBUS), which can effectively improve the prediction performance of the model. Hoyos-Osorio et al. \cite{41hoyos2021relevant} propose the RIUS under-sampling algorithm. They also combined RIUS and CBUS to obtain a new variant of the CRIUS under-sampling algorithm. The above algorithms use clustering algorithms to balance the class distribution of the data set, and do not or rarely consider the problem of class overlap. Different from the above algorithm, OBU \cite{42vuttipittayamongkol2018overlap} is a new class overlap undersampling algorithm. This algorithm uses soft clustering algorithm to obtain the membership degree, and uses the global algorithm to determine the overlap area and removes the instances. DBMUTE \cite{43bunkhumpornpat2017dbmute} uses a density clustering algorithm to detect overlapping areas, and finds and underinstances overlapping negative instances from the overlapping areas.
In addition to the above two types of algorithms, there are other types of undersampling algorithms. Kang Qi er al. \cite{44kang2016noise} provide an under-sampling algorithm called NUS. This algorithm combines the advantages of noise filters and under-sampling algorithm to achieve better classification performances. Liu et al. \cite{45liu2020design} use the concept of information granules in granular computing and used different granularity algorithms to construct information granules on most types of data sets to capture the data characteristics of this type. Then, according to the quality of the information granule, the information granule with the highest specificity value is selected to realize the under-sampling of the data set.

\section{Motivation}

Distance metrics is a common algorithm for calculating the differences between instances in the field of machine learning, which including Euclidean distance \cite{46li2009adaptive,47shih2004efficient}, Manhattan distance\cite{48chiu2009mobile}, cosine similarity\cite{49dehak2010front} , standard Euclidean distance and Mahalanobis distance\cite{50de2000mahalanobis,51mahalanobis1961experiments}. Among them, Euclidean distance is the most common similarity metrics widely used in the field of imbalanced data mining.
Tomek-Links algorithm is a data cleaning and undersampling technique from condensed nearest neighbor (CNN), which was proposed by Ivan Tomek \cite{24tomek1976two}. The algorithm uses the Euclidean distance to calculate the degree of similarity between instances, and improves the classification performance of the algorithm by eliminating the boundary majority instances on the imbalanced training datasets. Before using the Tomek-Links algorithm, we need to define the Tomek-Link pair concept, as shown in Definition 1.

\textbf{Definition 1 }(Tomek-Link pairs) If a pair of minimally Euclidean distanced neighbors $(\textbf{x}_i,\textbf{x}_j)$ with $\textbf{x}_i$ belonging to the positive class and $\textbf{x}_j$ belonging to the negative class. Let $d_{ED}(\textbf{x}_i,\textbf{x}_j)$ denote the Euclidean distance between $\textbf{x}_i$ and $\textbf{x}_j$. If there is no instance $\textbf{x}_p$ satisfies the following condition: $d_{ED}(\textbf{x}_i,\textbf{x}_p)<\textbf{x}_i,\textbf{x}_j)$ or $d_{ED}(\textbf{x}_j,\textbf{x}_p)<\textbf{x}_i,\textbf{x}_j)$ then the pair $(\textbf{x}_i,\textbf{x}_j)$ , was called as Tomek-Link pair.

After oversampling, many researchers often use the Tomek-Links algorithm as an auxiliary algorithm for class-overlap detection to prevent the generation of overlapping minority instances. Most studies have shown that the classification performance of the classifier can be improved by deleting the overlap majority instances, such as SMOTE + TL \cite{52batista2003balancing}. Unfortunately, the classic Tomek-Links undersampling algorithm has the following shortcomings: 

(1)When the number of minority instances is small, the overlapping instances detected by the classic Tomek-Links algorithm are not comprehensive, and it is easy to ignore the potential overlapping instances outside the boundary.

(2)The Tomek-Links algorithm uses the Euclidean distance to search the global information of the data set, and does not consider the influence of some features on the class-overlap.

(3)Since the Tomek-Link under-sampling algorithm uses Euclidean distance to calculate the similarity between instances, the calculated process doesn’t consider properties of data distribution, it would turn out that, doesn’t accurately reflect the practical distance arrangement.
When the data set is large, the complexity of Tomek-Links undersampling algorithm is usually prohibitively high. The unstable cuts-based instance selection (UCBSS) algorithm is a novel instance selection algorithm proposed by Xing et al. in 2018 (for details, please refer to \cite{53xing2018study}). The relevant definitions of unstable cut-points algorithm are as follows:

\textbf{Definition 2 }(Cut-point) \cite{54fayyad1992handling} Let $\bm{S}$ be a complete dataset. Assuming that the features are all numeric and the feature values are not repeated. When we sort the feature values in ascending order, the midpoint of two adjacent feature values is defined as a cut-point on the features, denoted by $T$.

\textbf{Definition 3 }(Unstable cut-point) \cite{54fayyad1992handling} Let $\bm{S}$ be a complete dataset,$\mathcal{X}$ be a feature set for inputs,$\bm{y}$ be a class-label set for outputs in the data set. Considering the problem of binary classification, i.e., the class-labels is $\bm{y}=\{y_0,y_1\}$.

Assuming that $\textbf{x}_i$ and $\textbf{x}_{i+1}$ are adjacent instances after certain feature sorted.

(1)If $\textbf{x}_i\in y_0$, $\textbf{x}_{i+1}\in y_1$ or $\textbf{x}_i\in y_1$, $\textbf{x}_{i+1}\in y_0$, the cut point between adjacent instances is called unstable cut point.

(2)If $\textbf{x}_i,\textbf{x}_{i+1} \in y_0$ or $\textbf{x}_i,\textbf{x}_{i+1} \in y_1$, the cut point between adjacent instances is called stable cut point.

We need to note that there may be special circumstances. Suppose there are three adjacent instances $\textbf{x}_i$, $\textbf{x}_{i+1}$ and $\textbf{x}_{i+2}$ after certain feature sorted, if $\textbf{x}_i\in y_0$, $\textbf{x}_{i+1},\textbf{x}_{i+2}\in y_1$ , at this time, the cut point $T_i$ between two adjacent instances of $\textbf{x}_i$ and $\textbf{x}_{i+1}$ is a unstable cut point, and the cut point $T_{i+1}$ between two adjacent instances of $\textbf{x}_{i+1}$ and $\textbf{x}_{i+2}$ is a stable cut point. Therefore, the instance $\textbf{x}_{i+1}$ in the middle is re-labeled by the stable cut point and the unstable cut point at the same time. 

The algorithm uses the imbalance ratio as the basis of instance selection, and realizes instance selection from data set by presetting a threshold, thereby reducing the algorithm's operation time and improving the classification performance of the classifier. They argument that, for the feature sequence, if the majority instance does not entail an unstable cut point with the minority instances as a neighbor, then it is redundant. During the instance compression process, such redundant instances will be deleted first. If we regard it as an under-sampling algorithm for imbalanced data, it can obtain relatively favorable classification performance. Unfortunately, the algorithm has some shortcomings that are difficult to make up in the instance compression process. For the imbalanced classifier, some of the shortcomings may be fatal.

(1)The UCBSS algorithm proposed in \cite{53xing2018study} can effectively protect minority instances, but the remaining majority instances and minority instances are considered as the nearest neighbors to each other. Therefore, there may be a large number of overlapping majority instances in the retained instances, which are not conducive to improve the classification performance of the classifier.

(2)In the instance selection process, it must be assumed that the value of each feature is completely different. In terms of real-world data sets, this assumption is rarely true, there are not many such data sets.

(3)In addition, the UCBSS algorithm assumes that each feature is independent of each other, but as we all know, there are often inextricable connections among features in the data set.

In order to mitigate the potential shortcomings of the above UCBSS and Tomek-Links algorithms, inspired by the core ideas of Tomek-Links algorithm and UCBSS algorithm, we use the multi-granularity learning framework in granular computing to fully consider the potential threat of local class-overlap majority instances, a new multi-granularity relabeled undersampling algorithm is proposed.

\section{The Proposed Algorithm}

Granular computing (GrC) has become an emerging concept and computing paradigm in information processing \cite{55wang2018granular}. Since granular computing was proposed, related models and theories have basically taken shape. Granular computing is not a specific model that can be used for classification or prediction, but a basic algorithm for simulating human thinking. It is a key step in data preprocessing and the underpinning of other modeling algorithms. Through granular computing, we can get the hidden information of the problem in different granularity subspaces, as well as different representations \cite{56wu2008granular}.
Multi-granularity data analysis is an important research content in the field of data mining. It conducts multi-angle and in-depth analysis and processing of data sets based on the idea of multi-granularity, and mines the potential information or knowledge representations in the data sets \cite{57wang2014granular}. Our proposed algorithm uses a multi-granularity learning framework, which can effectively mine the potential overlapping instances of the data set and improve the classification performance of the model.

Before proposing the MGRU algorithm, we need to use the following definitions:

\textbf{Definition 4 }(Local Granularity Subspaces) For binary classification problem, assuming that $\bm{S}=\{\textbf{x}_1,\textbf{x}_2,...,\textbf{x}_n\}$ is the training dataset, which contains $n$ instances, each instances has $m$ features $\mathcal{X}=(a_1,a_2,...,a_m)$ there is no feature vector with the same feature value and all zeros in the data set. Where $\bm{S}^{-\tau}$ can be obtained by deleting a certain feature $a_\tau$, $\tau =1,2,...,m$.This subset is called a local granularity subspace and can be written as a matrix form:

\begin{equation}
	\label{eq1}
\begin{array}{l}
 \begin{array}{*{20}c}
   {{\kern 1pt} {\kern 1pt} {\kern 1pt} {\kern 1pt} {\kern 1pt} {\kern 1pt} {\kern 1pt} {\kern 1pt} {\kern 1pt} {\kern 1pt} {\kern 1pt} {\kern 1pt} {\kern 1pt} {\kern 1pt} {\kern 1pt} {\kern 1pt} {\kern 1pt} {\kern 1pt} {\kern 1pt} {\kern 1pt} {\kern 1pt} {\kern 1pt} {\kern 1pt} {\kern 1pt} {\kern 1pt} {\kern 1pt} {\kern 1pt} {\kern 1pt} {\kern 1pt} {\kern 1pt} {\kern 1pt} {\kern 1pt} {\kern 1pt} {\kern 1pt} {\kern 1pt} {\kern 1pt} {\kern 1pt} {\kern 1pt} {\kern 1pt} {\kern 1pt} {\kern 1pt} {\kern 1pt} {\kern 1pt} {\kern 1pt} {\kern 1pt} {\kern 1pt} {\kern 1pt} {\kern 1pt} {\kern 1pt} {\kern 1pt} {\kern 1pt} {\kern 1pt} {\kern 1pt} {\kern 1pt} {\kern 1pt} {\kern 1pt} {\kern 1pt} {\kern 1pt} {\kern 1pt} {\kern 1pt} {\kern 1pt} {\kern 1pt} {\kern 1pt} {\kern 1pt} {\kern 1pt} {\kern 1pt} {\kern 1pt} {\kern 1pt} {\kern 1pt} {\kern 1pt} {\kern 1pt} {\kern 1pt}
 {\bf{a}}_1 {\kern 1pt} {\kern 1pt} } & {{\bf{a}}_2 } &  \cdots  & {{\bf{a}}_m } & {label}  \\
\end{array} \\ 
 {\cal S}^{ - \tau }  = \left[ {\begin{array}{*{20}c}
   {{\bf{x}}_1 }  \\
   {{\bf{x}}_2 }  \\
    \vdots   \\
   {{\bf{x}}_n }  \\
\end{array}} \right] = \left[ {\begin{array}{*{20}c}
   {a_{11} } & {a_{12} } &  \cdots  & {a_{1m} } & {{\kern 1pt} {\kern 1pt} {\kern 1pt} {\kern 1pt} {\kern 1pt} {\kern 1pt} 0}  \\
   {a_{21} } & {a_{22} } &  \cdots  & {a_{2m} } & {{\kern 1pt} {\kern 1pt} {\kern 1pt} {\kern 1pt} {\kern 1pt} {\kern 1pt} 1}  \\
    \vdots  &  \vdots  &  \ddots  &  \vdots  & {{\kern 1pt} {\kern 1pt} {\kern 1pt} {\kern 1pt} {\kern 1pt} {\kern 1pt}  \vdots }  \\
   {a_{n1} } & {a_{n2} } &  \cdots  & {a_{nm} } & {{\kern 1pt} {\kern 1pt} {\kern 1pt} {\kern 1pt} {\kern 1pt} {\kern 1pt} 0}  \\
\end{array}} \right]_{\left[ {n \times \left( {m - 1} \right)} \right] + {\cal Y}}  \\ 
 \end{array}
\end{equation}

\textbf{Theorem 1 }Let the Euclidean distance function be $f(\cdot ,\cdot)$,and two instances $\textbf{x}_i$ and $\textbf{x}_j$ in the data set $\bm{S}$. Suppose the distance is $f(\textbf{x}_i^{-\alpha},\textbf{x}_j^{-\alpha})$ when the feature $\alpha$ is deleted, and the distance is $f(\textbf{x}_i^{-\beta},\textbf{x}_j^{-\beta})$ when the feature $\beta$ is deleted. If the feature values satisfy $|a_{i\alpha}-a_{j\alpha}|\neq|a_{i\beta}-a_{j\beta}|$, then $f(\textbf{x}_i^{-\alpha},\textbf{x}_j^{-\alpha}\neq f(\textbf{x}_i^{-\beta},\textbf{x}_j^{-\beta})$.

Where the superscript $\alpha$ and $\beta$ represent the local granularity subspace formed by removing the $\alpha$-th or $\beta$-th features, and the subscript $i$ and $j$ represent the $i$-th or $j$-th instances. 

\textbf{Proof:} When feature $\alpha$ is deleted, the Euclidean distance between instances is $f(\textbf{x}_i^{-\alpha},\textbf{x}_j^{-\alpha})$, and when feature $\beta$ is deleted, the distance between instances is $f(\textbf{x}_i^{-\beta},\textbf{x}_j^{-\beta})$, and we need to prove that $f(\textbf{x}_i^{-\alpha},\textbf{x}_j^{-\alpha})=\sqrt{\sum_{l=1}^m(\textbf{x}_i^{-\alpha}-\textbf{x}_j^{-\alpha})^2}\neq f(\textbf{x}_i^{-\beta},\textbf{x}_j^{-\beta})=\sqrt{\sum_{l=1}^m(\textbf{x}_i^{-\beta}-\textbf{x}_j^{-\beta})^2},i,j=1,2,...,m$.

Then we square the two sides and expand to eliminate the same terms, which can be written as:$|(a_{i\alpha}-a_{j\alpha})^2|\neq|(a_{i\beta}-a_{j\beta})^2|$.

After removing the square, which can be written as $|a_{i\alpha}-a_{j\alpha}|\neq|a_{i\beta}-a_{j\beta}|$.

Therefore, if the feature values satisfy the inequality $|a_{i\alpha}-a_{j\alpha}|\neq|a_{i\beta}-a_{j\beta}|$, we have $f(\textbf{x}_i^{-\alpha},\textbf{x}_j^{-\alpha})\neq f(\textbf{x}_i^{-\beta},\textbf{x}_j^{-\beta})$.The conclusion holds.   

When the covariance matrix is the same and is a unit matrix or $\sum_i=\sum_j=1,i,j=1,2,...,m$, the Mahalanobis distance and the standardized Euclidean distance are equivalent to the classical Euclidean distance. Therefore, the result of Theorem 1 still applies.

For Mahalanobis distance and standardized Euclidean distance, if the covariance matrix is not a unit matrix and is not the same, or $\sum_i\neq\sum_j$, then the distance between the two instances may be equal. But in the arranging process, only a few instances have the same distance value. Therefore, this will not have a major impact on our overall ranking results.

In real-world data sets, there are very few cases where the features are exactly the same. Therefore, it is feasible to construct a local granularity space by removing a certain feature, and to mine the potential overlapping instances of the data set based on the data subset in the local granularity space. 

In Section 4.1, we will introduce the proposed MGRU undersampling algorithm in detail.

\subsection{MGRU Algorithm}

The MGRU algorithm we proposed is mainly divided into four Phases. The algorithm flowchart is shown as in Fig. 1. In the first phase, a local granularity subspace is formed by deleting some features. In the second stage, we calculate the Mahalanobis distance or standard Euclidean distance of all instances and rearrange them. In the third phase, we calculate the global relabeled index values of all instances to form a global relabeled index vector. In the fourth phase, we merge the global re-labeled index vector with the original data to form a re-labeled augmented data set   to carry out the under-sampling procedure.

\begin{figure}[!htbp]
	\centering
	\includegraphics[scale=1]{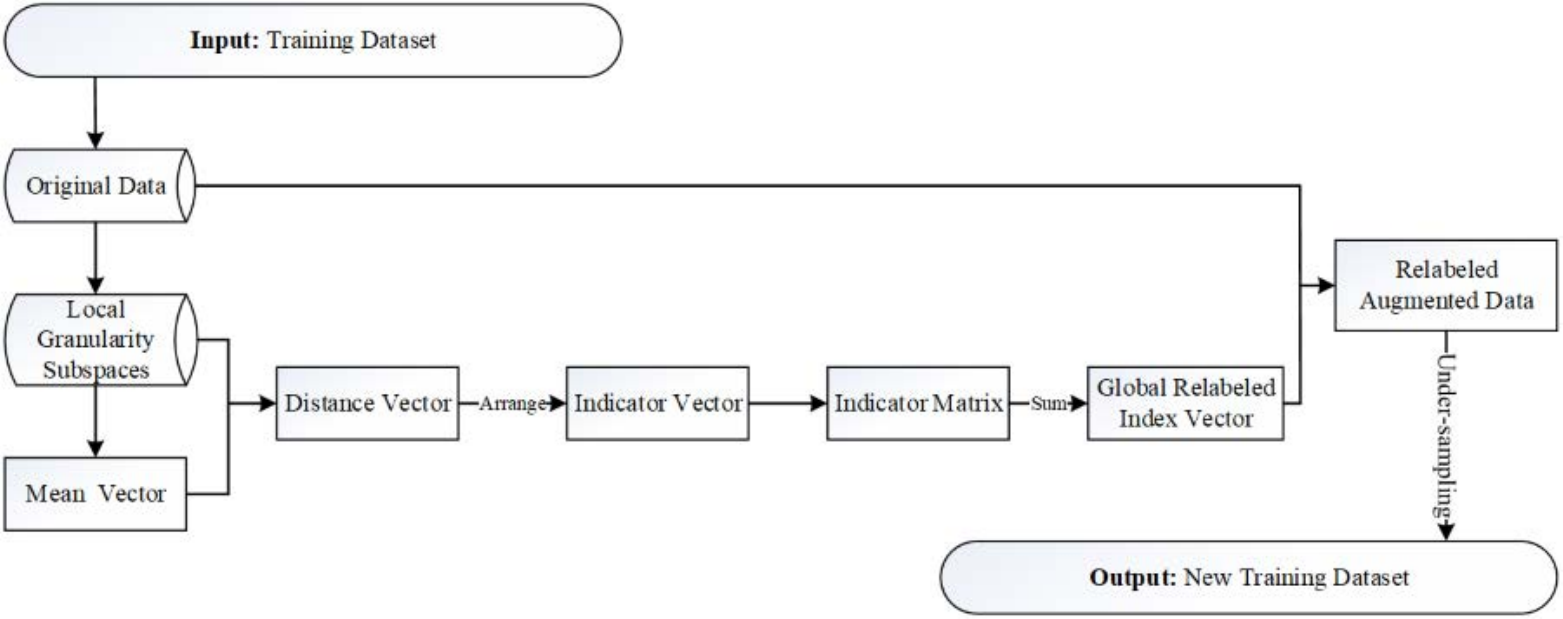}
	\caption{Flowchart of MGRU under-sampling algorithm}
	\label{fig1}
\end{figure}

\textbf{Phase 1 Generating the local granularity subspaces}

According to Theorem 1, it is not difficult to find that as long as the features we remove are not the same, the local granularity subspaces formed are not the same. Therefore, we can derive a variety of different combinations of local granularity subspaces. However, if the derived granularity subspaces are too rough, it is difficult for us to mine local overlapping instances. If the devised granularity subspaces are too fine, it may not be possible to mine more valuable local overlapping instances, but it will greatly increase our computation complexity. Therefore, in our experiment, we only delete each feature in the training data set one by one to formed a coarser local granularity subspace. Once the local granularity subspaces are constructed, we will enter the second phase.

\textbf{Phase 2 Calculating distances of all instances and getting the indicator vector}

At this phase, we obtain the relabeled index value of each instance through a series of calculation processes. The calculation process and related distance definitions are as follows.

Before introducing these measurement algorithms, we assume that there is a data set $\bm{S}$ with $n$ instances, each instance has $m$ features, and $\mu =(\mu_1,\mu_2,...,\mu_m)$ is the mean vector of all instances.

\textbf{Definition 5 }(Standard Euclidean distance,$d_{ED}$): If $\textbf{x}_i$ and $\textbf{x}_j$ are instances in the data set $\bm{S}$, the standardized Euclidean distance $d_{ED}$ is defined as follows:

\begin{equation}
	\label{eq2}
	d_{ED}=\sqrt{\sum_{k=1}^m(\frac{\textbf{x}_{ik}-\textbf{x}_{jk}}{S_k})^2},i,j=1,2,...,n   
\end{equation}

where $s_k$ is the standard deviation of the instances’ k-th component. If the reciprocal of the standard deviation is regarded as a weight, this formula can be regarded as a weighted Euclidean distance.

Indian statistician P. C. Mahalanobis \cite{51mahalanobis1961experiments} first proposed the Mahalanobis distance, introducing a covariance matrix to calculate the similarity between instances in the data set. 

\textbf{Definition 6 }(Mahalanobis distance,$d_{MD}$) \cite{51mahalanobis1961experiments}: If $\textbf{x}_i$ is a instance in the data set $\bm{S}$, and the covariance matrix for data set $\bm{S}$ is $\sum=(\sigma_{ij})$, then the Mahalanobis distance of the instance $\textbf{x}_i$ is defined as:

\begin{equation}
	\label{eq3}
	d_{MD}=\sqrt{(\textbf{x}_i-\mu)^T\Sigma ^{-1}(\textbf{x}_i-\mu)},i=1,2,...,n   
\end{equation}

Compared with Euclidean distance, Mahalanobis distance and standard Euclidean distance consider the relationship between instance, and can more accurately show the structural differences between instances. Therefore, in this study, the Mahalanobis distance or the standard Euclidean distance are used to measure the similarity between instances in the local granularity subspace.

Next, we introduce the detailed calculation process. This phase is mainly divided into two steps. The first step is to use the Mahalanobis distance or standardized Euclidean distance to calculate the similarity between the instances. The second step is to re-label the instances’ indexes according to the similarity, and obtain the indicator vector. More precise, it is divided into two step:

\textbf{Step 1 }We calculate the distance between instances and the instances’ mean according to Definition 5 or Definition 6, for the local granularity subspaces, we rearrange all instances in ascending order according to their distances. When we choose the Mahalanobis distance for arranging, we can get the multi-granularity Mahalanobis distance relabeled under-sampling algorithm (MGRU-MD), and when we use the standardized Euclidean distance, we can obtain the multi-granularity standardized Euclidean distance relabeling under-sampling algorithm (MGRU-SED). 

Note that these two under-sampling algorithms only differ in the distance calculation steps, and the calculation process is the same in the remaining phase, so in the remaining steps, we will uniformly introduce the overall calculation process.

\textbf{Step 2 }After all the instances are rearranged, it will enter the re-labeled instance phase. The detailed calculation process is as follows:

We sort the results according to distance and search for Tomek-Link pairs in the local granularity subspaces. Then, we re-labeled the majority instances in the Tomek-Link pair, and calculate the indicator value of each instance on each subspace, which is represented by $\lambda_j^{-\tau}$, where the superscript $-\tau$ signifies the local granularity subspace formed by removing the $\tau$-th $(\tau=1,2,...,m)$ feature, and the subscript $j$  denotes the $j$-th (j=1,2,...,n) instances. When we have re-labeled, we combine the re-labeled values of all instances to form the corresponding indicator vector, which is represented by $\delta^{-\tau}=(\lambda_1^{-\tau},\lambda_2^{-\tau},...,\lambda_n^{-\tau})$.

Note that during the re-labeled process, the indicator value may be repeatedly obtained. If the instance has repeated indicator values, we will only record the largest indicator value once.

\textbf{Phase 3 Constructing indicator matrix and calculating global relabeled index values for all instances }

We merge all indicator vectors to form a corresponding indicator matrix, denoted by $\textbf{M}=[\delta^{-1},\delta^{-2},\\...,\delta^{-m}]^T$. And use formula (4) to calculate the global relabeled index value of the instances.

\begin{equation}
	\label{eq4}
	K(\textbf{x}_j)=\sum_{\tau=1}^m\lambda_j^{-\tau}  
\end{equation}

In short, the global relabeled index value corresponding to each instance can be calculated as the sum of each row in the indicator matrix. The vector formed by all re-labeled index values is called the global re-labeled index vector, denoted by $\Phi=(K(x_1),K(x_2),...,K(x_n))^T$.

\textbf{Phase 4 Data fusion and undersampling procedure}

At this stage, we merge the re-labeled index vectors of the obtained instance with the original data set to form a new re-labeled augmented data set. More specifically, assuming that the original training data set is $\bm{S}$, we incorporate the re-labeled index vector formed in the third stage into the training set to form a re-labeled augmented training data set $\bm{S^+}=[\bm{S},\Phi]$.

When the re-labeled augmented training data set is constructed, our preprocessing is basically over. For the under-sampling algorithm, we only need to select certain index value in the global re-labeled index vector as the threshold, and then delete instances less than or equal to the threshold to complete the under-sampling step. Since there is no clear guidance algorithm for how to determine the threshold at present, in the experimental part, we use the greedy search algorithm to select the sub-optimal threshold. 

\subsection{Global relabeled index value selection}

The global re-labeled index value indicates the number of times that majority instances form Tomek-Links pairs in the local granularity subspaces. The more times it is re-labeled, it indicates that the instance is more likely to be an overlapping instance. For our under-sampling algorithm, the choice of global re-labeled index value is very important. If the selected value is large, overlapping instances may be deleted incompletely, otherwise, it may cause excessive elimination. Therefore, in our experiments, under the premise of ensuring that the data set partition remains unchanged, the idea of greedy search is used to traverse each global relabeled index value $K$ to obtain the best score and the best global re-labeled index value of the model.

For the convenience of description, we might as well assume that $F(K),K=1,2,...,m$ is the score of the classifier on a certain global relabeled index value.

If $F(K)>F(K-1)$, we save the score of the classifier when the value of $F(K)$ , we save the score of the classifier when the value of $F(K-1)$.In the next search, the global re-labeled index value $F(K-1)$ will not be selected.

If $F(K)\leq F(K-1)$, we will update the classifier score and at the same time update the optimal relabeled index value.
After traversing all the relabeled index values, there will be no search, and the model outputs the optimal relabeled index value and the final score of the classifier. The pseudocode for our proposed algorithm is shown in Algorithm 1. 

\begin{algorithm}
	\renewcommand{\algorithmicrequire}{\textbf{Input:}}
	\renewcommand{\algorithmicensure}{\textbf{Output:}}
	\caption{Mulit-granularity Re-labeled under-sampling algorithm}
	\label{alg1}
	\begin{algorithmic}[1]
		\REQUIRE training set $\bm{S}$, labels set $\bm{y}$.
		\ENSURE data set $\bm{S'}$ and $\bm{y'}$ after the under-sampling
		\STATE  Initialize relabeled index vector $\Phi=\{0,0,...,0\}$
		\STATE  \textbf{for} each feature $a_\tau$ in $\bm{S}$ do
		\STATE  Initialize indicator vector $\delta^{-\tau}=(\lambda_1^{-\tau},\lambda_2^{-\tau},...,\lambda_n^{-\tau},)^T$
		\STATE $\bm{S^{-\tau}}=\bm{S}-a_\tau$	
		\STATE Set $\mu=mean(\bm{S^{-\tau}})$
		\STATE Set $d_j$ = Mahalanobis distance or standardized Euclidean distance between $\textbf{x}_j$ and $\mu$
		\STATE QuickSort($d_j$)
		\STATE \textbf{if} $d^{-\tau}(\textbf{x}_i,\textbf{x}_j)=min$ \textbf{then}:
		\STATE \qquad\textbf{if} $\textbf{x}_j=majority$ \textbf{then}:
		\STATE \qquad\qquad$\lambda_j^{-\tau}$ =1;
		\STATE \qquad\textbf{end if}
		\STATE \textbf{end if}
		\STATE \qquad$K(\textbf{x}_j)+=\lambda_j^{-\tau}$
		\STATE \textbf{end for}
		\STATE set threshold value $K$
		\STATE \textbf{for} every instance $\textbf{x}_j$ in $\bm{S}$ \textbf{do}
		\STATE \qquad\textbf{if} $K(\textbf{x}_j)\leq K$ \textbf{then}
		\STATE \qquad\qquad delete $\textbf{x}_j$
		\STATE \qquad\textbf{end if}
		\STATE \textbf{end for}
	\end{algorithmic}
\end{algorithm}

We use ten-fold cross-validation in the experiments to verify the stability of the model. In order to better describe the classification performance of the model on a certain relabeled index value, we choose the same index value on different fold data sets. Note that in the training process, there may be cases where the relabeled index value of each fold does not exist, of course, the probability of this situation is very small. When the data is concentrated and the amount of training data is large, this situation can be ignored. Therefore, if this happens to a certain fold, we choose the nearest larger relabeled index value for calculation.

Although this algorithm can obtain the best performance and the best index value of the classifier, the amount of calculation is relatively large, and each index value needs to be traversed. Therefore, in the next research, we will further explore the adaptive index value selection algorithm, use information from the data set to calculate the optimal index value, and shorten the calculation time. 

\subsection{Algorithm Complexity}

In this section, we derive the computational complexity of the proposed MGRU algorithm. Let the data set be $\bm{S}$, where the number of instances is $n$, the number of features is $m$, and the number of categories is $c$. To analyze the computational complexity for the MGRU algorithm, first, we need to delete the features in the data set one by one to form a de-featured subspace. This process is simple and the calculation complexity is low, so we ignore it. Then, for the Mahalanobis distance, the time complexity of calculating the instance mean $\mu$ is $O(nc)$. For the instance covariance matrix, each elements requires $n$ times of multiplication and $m$ times of addition, so its computational complexity is $O(n^2m)$. Then, we need to consider the complexity of sorting process. In this paper, quick sort is our choice. At this time, it is well known that the best sorting complexity is $O(nlogn)$, and the worst sorting complexity is $O(n^2)$. Finally, we need to calculate the relabeled index value of the instance, for this aim, we only need to calculate the sum of $n$ times, therefore, the time complexity is $O(n)$. In our calculation process, we need to traverse the results corresponding to each threshold in order to obtain the sub-optimal results. Therefore, we need to calculate the whole process $m$ times. Therefore, the overall complexity of the MGRU algorithm is $O(n^2m\times m)$.

Based on the above analysis, the dominant complexity of the MGRU algorithm is mainly affected by the Mahalanobis distance calculation process and sorting algorithm. Of course, if we need relatively low time complexity, we can use the squared Mahalanobis distance for sorting, or we can choose the sorting algorithm with lower time complexity according to the size of the instance set.

\section{Experiments framework}

In this paper, we use Python 3.8 to implement algorithm simulation. We compared our proposed algorithm with the state-of-the-art algorithm on 46 real-world data sets to verify the effectiveness of the proposed algorithm. The hardware and software configuration of the experiment computer are given in Table 1.

\begin{table}[!htbp]
	\centering
	\caption{The hardware and software configuration of the experiment computer}
	\label{tb1}
	\begin{tabular}{ccc}
		\hline	
\qquad	& Items & Configuration\\ \hline
\multirow{2}{*}{Hardware configuration} &CPU	&Intel(R) Core(TM) i7-6700\\
&Memory	&16GB\\
\multirow{3}{*}{Software configuration} &Hard disk	&Solid state disk(1TB)\\
&Operating system	&Windows 10 Professional\\
&Interpreter	&Python 3.8\\                                     
 \hline
	\end{tabular}
\end{table}

The rest of this section mainly introduces our experimental setup and the data set used. We introduce the experimental setup and model in Section 5.1. The 20 real-world highly imbalanced data sets used are briefly introduced in Section 5.2.

\subsection{Setup}

In our experiments, all experiments use 10-fold cross-validation to verify the stability of the model. The experimental results show in this paper are the average of the 10-fold cross-validation results. Note that the unstable cuts-based instance selection (UCBSS) algorithm is used to label both the majority class and the minority class instances. Therefore, when they compress instances, they keep instances closer to the minority instances. In order to adapt the algorithm to the problem of class overlap, we make minor changes to UCBSS on the basis of the original algorithm. The changes are as follows: 

(1)In the UCBSS marking process, we also only mark the majority instances near the unstable cut point, and not mark the minority instances.

(2)When there are the same values in the features, we arrange the instances in a random order.

In order to verify the effectiveness of the proposed algorithm, we compared the MGRU-MD and MGRU-SED class-overlap undersampling algorithms with seven algorithms including UCBSS on 20 real-world highly imbalanced data sets. These algorithms include 5 resampling algorithms and 2 ensemble learning algorithms. 

\textbf{Null} is our baseline comparison method without any pre-processing of the training set.

\textbf{SMOTE} \cite{58chawla2002smote} is a well-known oversampling method for imbalanced data pre-processing. In the calculation process, it randomly selects minority instances and uses linear interpolation to synthesize more minority instances. 

\textbf{Tomek-Links} \cite{24tomek1976two} algorithm is a class-overlap under-sampling algorithm proposed by Tomek. The algorithm considers the majority instances in the Tomek-Link pair to be overlapping instances in the data set. Therefore, in the under-sampling process, majority instances in the Tomek-Link pair are deleted.

\textbf{NB-TL} \cite{37vuttipittayamongkol2020neighbourhood} and \textbf{NB-Comm} \cite{37vuttipittayamongkol2020neighbourhood} are under-sampling frameworks for processing overlapping instances in data sets, proposed by Vuttipittayamongkol et al. Under this framework, they proposed four k-nearest neighbor class-overlap under-sampling algorithms with different criteria. The experimental results show that the performance difference between $k=3$ and $k=5$ is not obvious, and the performance of NB-TL and NB-Comm is better than the other two algorithms. hence, below we only compare our two algorithms with NB-TL and NB-Comm with $k=5$.
\textbf{RUSBoost} \cite{59seiffert2009rusboost} is a classic ensemble learning algorithm that uses undersampling as a preprocessing technique. In our experiments, we use the optimal parameters in \cite{59seiffert2009rusboost} to conduct experiments. Since RUS is a heuristic algorithm, it has strong randomness in the calculation process, so its performance is not stable.
\textbf{SPE} \cite{60liu2020self} (Self-paced ensemble learning) is a new ensemble learning framework proposed by Liu et al. using self-paced learning. Since they use self-paced learning to gradually eliminate majority instances in the ensemble process, they can gradually balance the training set of subsequent classifiers. Therefore, it is more in line with the requirements of our comparison algorithm.
The above algorithms are all preprocessing algorithms and ensemble learning frameworks independent of the classifier. Therefore, in our experiment, we select Classification And Regression Tree (CART), Support Vector Machines (SVM) and Gradient Boosting Decision Tree (GBDT) three classifiers as the base classifiers in the experiment. In order to better highlight the performance of the preprocessing algorithm, in our experiments, these three classifiers all use the default parameters in sklearn for experimentation, because this can keep the parameters unchanged, thus highlighting the superiority of the preprocessing or integration framework.

\subsection{Datasets}

In our experiments, we collected 20 imbalanced binary data sets from the KEEL database. The basic information of the data set is shown in Table 2 (in order of imbalance ratio from small to large). The imbalance ratios (IR) of all instances are ranged from 1.87 to 129.44. The traditional sampling algorithms maintain the balance between the negative and positive instances in the training set by calculating, thereby improving the classification accuracy of the classification model. The imbalance ratio is an important indicator to measure the imbalanced distribution on a data set, and its expression is shown in Eq.(5):

\begin{equation}
	\label{eq5}
	IR=\frac{|N_{maj}|}{|N_{min}|}
\end{equation}

where $|N_{maj}|$ represents the number of instances in the majority (negative) class, and $|N_{min}|$ represents the number of instances in the minority (positive) class.

\begin{table}[!htbp]
	\centering
	\caption{Basic information of the data set}
	\label{tb2}
	\begin{tabular}{cccccc}
		\hline	
Datasets	&feature	&Instance	&Minority	&class &IR \\ 
\hline
pima	     &8	&768	&268	&2	&1.87\\
glass0	&9	&214	&70	&2	&2.06\\
haberman	&3	&306	&81	&2	&2.78\\
vehicle1	&18	&846	&217	&2	&2.9\\
vehicle0	&18	&846	&199	&2	&3.25\\
ecoli1	&7	&336	&77	&2	&3.36\\
ecoli2	&7	&336	&52	&2	&5.46\\
yeast3	&8	&1484	&163	&2	&8.1\\
page-blocks0	&10	&5472	&559	&2	&8.79\\
yeast-2\_vs\_4	&8	&514	&51	&2	&9.08\\
abalone9-18	&8	&731	&42	&2	&16.4\\
winequality-red-4	&11	&1599	&53	&2	&29.17\\
yeast5	&8	&1484	&44	&2	&32.73\\
winequality-red-8\_vs\_6	&11	&656	&18	&2	&35.44\\
abalone-17\_vs\_7-8-9-10	&8	&2338	&58	&2	&39.31\\
abalone-21\_vs\_8	8	&581	&14	&2	&40.5\\
winequality-white-3\_vs\_7	&11	&900	&20	&2	&44\\
abalone-20\_vs\_8-9-10	&8	&1916	&26	&2	&72.69\\
poker-8-9\_vs\_5	&10	&2075	&25	&2	&82\\
abalone19	&8	&4174	&32	&2	&129.44\\                                  
 \hline
	\end{tabular}
\end{table}

\section{Results and analysis}

In this section, we will conduct a complete experimental analysis and verify the effectiveness of the proposed algorithm through three experiments:

(1)First, we use the class-overlap complexity analysis proposed by Pascual-Triana et al. \cite{61pascual2021revisiting} to verify that the algorithm can effectively reduce the complexity of the data set. For multiple comparisons we use the Friedman test \cite{62garcia2009study,63garcia2010advanced} to detect statistical differences among a group of results, and then utilize Holm post-hoc test \cite{63garcia2010advanced} to find which algorithms are distinctive among a   comparison (Section 6.1).

(2)Secondly, under the two evaluation metrics of the Area Under the ROC Curve (AUC) and the Area Under the PR Curve (auPR), we compare and analyze the proposed MGRU algorithm with other state-of-the-art algorithms to verify the effectiveness of the model, and use the same algorithms in section 6.1 to verify the statistical significance of the model (Section 6.2).

(3)Finally, we will summarize and analyze the experimental results and discuss them in Section 6.3.

Note that the complexity measurement in the experiment is performed on the training set. Therefore, when we choose the optimal AUC value, we also use   to measure the complexity of the training set after sampling.

\subsection{Class-overlap complexity measurement after sampling}

The data complexity measurement is an algorithm to evaluate the problem of feature overlap, linear separability, and so on. On the basis of the complexity measure proposed by Ho et al. \cite{64ho2002complexity}, many researchers have further studied the problems of overlap, separability, geometry, topology and density. Pascual-Triana et al. \cite{61pascual2021revisiting} proposed two new algorithms for the overlap complexity measurement of datasets called $ONB_{tot}$ and $ONB_{avg}$.The experimental results in \cite{61pascual2021revisiting} show that for the overlapped imbalanced data set, whether using Euclidean distance or Manhattan distance, the $ONB_{avg}$ complexity metric can effectively evaluate the degree of overlap of the data set. Since the used datasets is characterized by continuous values, the use of Manhattan distance is easy to distort the data. Therefore, in our experiment, the Euclidean distance is used as the metric function. The calculation process of $ONB_{avg}$ is shown in Eq. (6). If the value of $ONB_{avg}$ is higher, it means that the degree of overlap of the data set is higher, and the data is more complicated.

\begin{equation}
	\label{eq6}
ONB_{avg}  = \frac{{\sum {_{i = 1}^k {\textstyle{{b_i } \over {n_i }}}} }}{k}	
\end{equation}

where $b_i$ is the number of balls for class i,$n_i$ is the number of elements of said class and $k$ is the number of classes.

$ONB_{avg}$ is a conservative class-overlap complexity metric. If we only delete the majority instances covered by the overlapping hyperspheres, the problem of overlapping data sets cannot be eliminated. At this time, since the mistaken deletion of majority instances, the number of denominators of $ONB_{avg}$ decreases, but the numerator does not change at all, and the obtained $ONB_{avg}$ value does not decrease but increases. According to the calculation process of $ONB_{avg}$, if the number of overlapping instances is not changed during the sampling process, but the number of instances is increased, the value of $ONB_{avg}$ will also be significantly reduced. Since in the SMOTE and the other oversampling technique of the interpolation algorithm, the nearest neighbors of the anchored instances are used in the training set to synthesize new training instances. When the imbalance ratio of the data set is large, there are more pseudo instances added, which will dilute the total number of samples in the overlapping area of the data set. Even if we do not exclude the majority instances in any overlapping area, $ONB_{avg}$ will decrease. Therefore, in the experiment, we did not consider comparing SMOTE with the under-sampling algorithm. In addition, the two ensemble learning algorithms of SPE and RUSBoost differ in the number of instances of majority classes eliminated by the base classifier during the training process. Therefore, we cannot accurately calculate the class-overlap complexity corresponding to the training set.

Based on above discussion, we only use the new proposed overlap and undersampling algorithms for comparison, including Tomek-Links, NB-TL, NB-Comm and UCBSS. The description and parameters of the algorithm are as described in section 5.1. In order to be consistent with the subsequent experiments, we use ten-fold cross-validation. We use $ONB_{avg}$ to measure the complexity of the training set after sampling on the corresponding training set. Then we calculate the mean value of all measurement results and use it as the final experimental result. The entire complexity results of this section are listed in Table 3. For the convenience of reading, we highlight the lowest $ONB_{avg}$ value corresponding to each data set in bold. The average $ONB_{avg}$ complexity of each algorithm is shown in Fig. 2. The average Friedman rankings of and APVs using Holm’s post-hoc test in $ONB_{avg}$ are shown in Table 4.

\begin{table}[!htbp]
	\centering
	\label{tb3}
	\caption{The class overlap complexity after sampling using  as metric.}
	\begin{tabular}{cccccccc}
		\hline
		dataset	&Null	&Tomek-Links	&NB-TL	&NB-Comm	&UCBSS	&MGRU-MD	&MGRU-SED \\ \hline
pima	     &0.4129	&0.3603	&0.3083	&0.3105	&0.2846	&0.2934	&0.2641\\
glass0	&0.2657	&0.2462	&0.2709	&0.4702	&0.3499	&0.2859	&0.2808\\
haberman	&0.3758	&0.3531	&0.3255	&0.3146	&0.3052	&0.3124	&0.3317\\
vehicle1	&0.1455	&0.1436	&0.1418	&0.1522	&0.1124	&0.1325	&0.1401\\
vehicle0	 &0.1455	&0.1463	&0.1681	&0.1507	&0.1357	&0.1084	&0.1426\\
ecoli1	 &0.2277	&0.1744	&0.1462	&0.3653	&0.2875	&0.1957	&0.1989\\
ecoli2	 &0.1608	&0.1663	&0.1921	&0.3211	&0.1729	&0.1428	&0.1547\\
yeast3	 &0.2141	&0.2684	&0.2382	&0.2059	&0.2238	&0.1843	&0.1965\\
page-blocks0	&0.2623	&0.2608	&0.2231	&0.2584	&0.2408	&0.2438	&0.2362\\
yeast-2\_vs\_4	&0.2162	&0.1938	&0.1471	&0.2849	&0.2712	&0.2959	&0.1227\\
abalone9-18	&0.4154	&0.4136	&0.3362	&0.3814	&0.3744	&0.3268	&0.3105\\
winequality-red-4	&0.4025	&0.3854	&0.4107	&0.3956	&0.3658	&0.3121	&0.3415\\
yeast5	&0.2315	&0.2031	&0.2416	&0.2635	&0.2318	&0.2033	&0.1988\\
winequality-red-8\_vs\_6	&0.4961	&0.4494	&0.4188	&0.4264	&0.3898	&0.3631	&0.3709\\
abalone-17\_vs\_7-8-9-10	&0.3468	&0.3536	&0.3309	&0.3605	&0.3102	&0.3264	&0.2918\\
abalone-21\_vs\_8	     &0.3417	&0.3415	&0.3409	&0.3783	&0.3961	&0.3489	&0.3376\\
winequality-white-3\_vs\_7	&0.3934	&0.3928	&0.3814	&0.3655	&0.3482	&0.3356	&0.3279\\
abalone-20\_vs\_8-9-10	&0.4205	&0.3826	&0.4108	&0.4622	&0.3805	&0.3527	&0.3699\\
poker-8-9\_vs\_5	     &0.4848	&0.4803	&0.4626	&0.4915	&0.4822	&0.4704	&0.4655\\
abalone19	&0.4903	     &0.4529	&0.4684	&0.4745	&0.4328	&0.3943	&0.3933\\
 \hline            
	\end{tabular}
\end{table}

\begin{figure}[!htbp]
	\centering
	\includegraphics[scale=1]{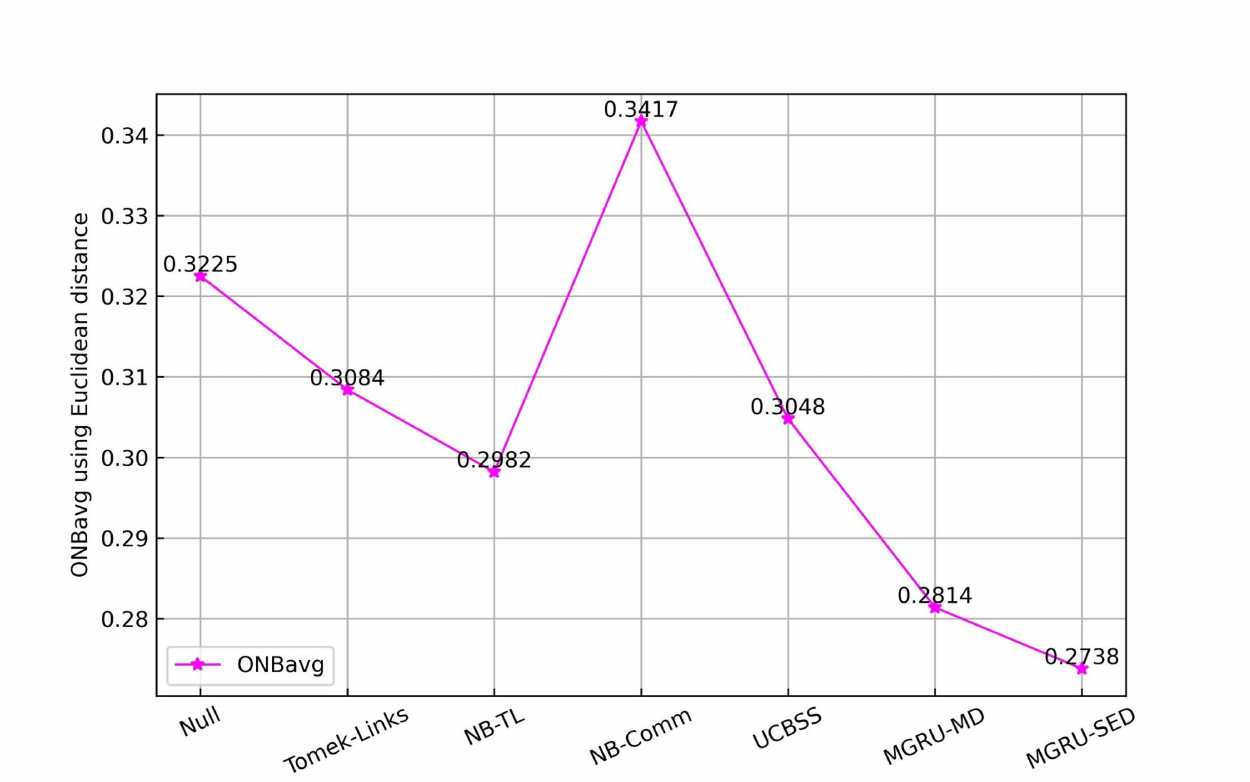}
	\caption{Average value of   in the experimental datasets.}
	\label{fig2}
\end{figure}

\begin{table}[!htbp]
	\centering
	\label{tb4}
	\caption{The class overlap complexity after sampling using  as metric.}
	\begin{tabular}{ccc}
		\hline
		Result (ONBavg)	&Friedman ranking	&APVs \\ \hline
MGRU-MD	&2.050	&-\\
MGRU-SED	&2.600	&0.420752\\
UCBSS	&3.700	&0.031440\\
NB-TL	&4.000	&0.012931\\
Tomek-Links	&4.500	&0.001341\\
Null	&5.400	&0.000005\\
NB-Comm	&5.750	&0\\
 \hline            
	\end{tabular}
\end{table}

According to the results in Table 3, it can be observed that on most data sets, compared with the overlapping complexity without sampling (null in the table represents the overlapping complexity of data without sampling), our proposed algorithm can reduce the complexity of training data sets. Compared with other state-of-the-arts class-overlap undersampling algorithms, MGRU-MD and MGRU-SED algorithms do not significantly improve the complexity of training data on glass0, haberman, vehicle1, ecoli1, page-block0 and poker-8-9\_vs\_5, but the difference is not obvious.

Figure 2 shows the average value of the class-overlap complexity $ONB_{avg}$ after sampling for each algorithm on all data sets. We can observe that the two algorithms we proposed have better overall performance in reducing the $ONB_{avg}$ overlap-complexity. NB-Comm is a new class-overlap undersampling algorithm using Tomek-Links. In the process of finding overlapping instances, the NB-Comm believes that the common majority instances of all minority instances are overlapping instances. For a highly imbalanced data set, there are not many minority instances in the data set. Therefore, this algorithm is too conservative, and it is likely to ignore the potentially overlapping majority instances of anchor instances. Therefore, this algorithm may render incomplete deletion, which causes the class-overlap complexity $ONB_{avg}$ to increase after majority samples are deleted.

According to the results in Table 4, we observe that MGRU-MD and MGRU-SED achieve the best Friedman rankings. Therefore, we believe that MGRU-MD and MGRU-SED are the best algorithms, and MGRU-SED has more potential advantages. In addition, compared with other class-overlap under-sampling algorithms, all APVs values calculated using Holm’s post-hoc test are lower than the significant level (e.g.$\alpha=0.05$ ). Therefore, the null hypothesis of equality is rejected in all cases. At the same time, the APVs of MGRU-MD and MGRU-SED are greater than 0.05. We believe that the performance difference between MGRU-MD and MGRU-SED is not significant in reducing class overlap complexity. Therefore, this also supports the conclusion that our proposed algorithm is superior to other algorithms. Next, we will use the two metrics of AUC and auPR to verify the effectiveness of the algorithms.

\subsection{Compared with state-of-the-art algorithms}

In this section, in order to further illustrate the effectiveness of our algorithm, we use two evaluation metrics, AUC and auPR, to evaluate the performance difference between MGRU-MD and MGRU-SED and other state-of-the-arts models. In the experiment, we use three advanced classifiers: CART, SVM and GBDT as base classifiers. And we will compare the proposed MGRU-MD and MGRU-SED with the seven state-of-the-arts algorithms ,i.e., SMOTE, Tomek-Links, NB-TL, NB-Comm, UCBSS, SPE and RUSBoost. SPE is a state-of-the-arts ensemble learning framework proposed by Liu et al. \cite{60liu2020self} in ICDE 2020. This algorithm eliminates majority instances one by one by using self-paced learning.  As an under-sampling algorithm in the ensemble framework, self-paced learning can effectively reduce the imbalance ratio of the training data set. Of course, we regard it as a state-of-the-arts ensemble learning algorithm combined with under-sampling algorithms. Note that we still use the mean of the 10-fold cross-validation as the final result of the model. The specific description of these algorithms and the selection of parameters have been introduced in detail in Section 5.1.

In order to better display the experimental results, we arrange the data set in ascending order of imbalance ratio, and the best results are highlighted in bold. Table 5-6 show the classification performance of our three classifiers, CART, SVM, and GBDT, and seven comparison algorithms under the two evaluation metrics of AUC and auPR. Fig. 3-5 show the average performance of the three classifiers on all data sets, where the left ordinate represents AUC, the right ordinate represents auPR, and the abscissa represents the corresponding algorithm. In addition, in order to evaluate the performance of the algorithm with statistically significant differences, the results of the non-parametric tests are given in Table 7. The second column is the Friedman ranking results of each algorithm on all data sets. In the third column, we use Friedman's test and Holm’s post-hoc test to calculate the APVs between each algorithm. Note that in Table 8, we have retained six decimal places. Therefore, when the value of APVs in the table is 0, it just means that the value is very small, approximately 0.

\begin{table}[!htbp]
	\centering
	\label{tb5}
	\caption{AUC of all algorithms in the collected datasets}
	\resizebox{\textwidth}{110mm}{
	\begin{tabular}{cccccccccccc}
		\hline
dataset	&models	&Null	&SMOTE	&TL	&NB-TL	&NB-Comm	&UCBSS	&SPE	&RUSBoost	&MGRU-MD	&MGRU-SED \\ \hline
\multirow{3}{*}{pima}	&CART	      &0.6626	&0.6782	&0.6896	&0.7141	&0.6881	&0.7217	&0.7175	&0.6992	&0.6983	&0.7312\\
	&SVM	&0.8156	&0.8139	&0.8099	&0.8115	&0.7999	&0.8199	&0.8117	&0.7668	&0.8146	&0.8252\\
	&GBDT	&0.8192	&0.8231	&0.8236	&0.8143	&0.8015	&0.8124	&0.8243	&0.8271	&0.8171	&0.8334\\\hline
\multirow{3}{*}{glass0}	&CART	&0.7729	&0.8052	&0.8157	&0.7881	&0.6924	&0.8164	&0.8301	&0.8284	&0.8212	&0.8067\\
	                     &SVM	&0.7186	&0.3276	&0.8177	&0.7866	&0.5672	&0.8305	&0.5845	&0.2966	&0.8386	&0.8114\\
	                     &GBDT	&0.9277	&0.9265	&0.9279	&0.9028	&0.7686	&0.9282	&0.9224	&0.9214	&0.9345	&0.9223\\\hline
\multirow{3}{*}{haberman}	&CART	&0.5573	&0.5875	&0.5249	&0.5711	&0.5719	&0.6103	&0.5738	&0.6014	&0.6275	&0.5719\\
	                     &SVM	&0.6438	&0.7061	&0.7116	&0.7039	&0.5634	&0.6820	&0.6806	&0.3265	&0.7223	&0.7154\\
	                     &GBDT	&0.6431	&0.6339	&0.6498	&0.6427	&0.6448	&0.6551	&0.6572	&0.6258	&0.6747	&0.6551\\\hline
\multirow{3}{*}{vehicle1}	&CART	&0.6889	&0.7051	&0.6856	&0.7293	&0.6784	&0.7258	&0.7884	&0.7643	&0.7368	&0.7152\\
	                     & SVM	&0.8389	&0.7439	&0.8021	&0.7335	&0.7365	&0.8445	&0.7163	&0.3731	&0.8459	&0.8406\\
	                     &GBDT	&0.8668	&0.8754	&0.8708	&0.8539	&0.8333	&0.8885	&0.8662	&0.8647	&0.8797	&0.8702\\\hline
\multirow{3}{*}{vehicle0}	&CART	&0.9073	&0.9297	&0.9084	&0.9222	&0.7668	&0.9385	&0.9523	&0.9430	&0.9443	&0.9351\\
	                     &SVM	&0.9117	&0.9595	&0.8993	&0.8741	&0.9224	&0.9675	&0.9088	&0.8618	&0.9698	&0.9299\\
	                     &GBDT	&0.9921	&0.9941	&0.9912	&0.9913	&0.9221	&0.9941	&0.9928	&0.9908	&0.9928	&0.9931\\\hline
\multirow{3}{*}{ecoli1}	&CART	&0.8489	&0.8485	&0.8758	&0.8862	&0.8576	&0.8758	&0.8297	&0.8108	&0.8921	&0.8791\\
	                     & SVM	&0.9122	&0.9227	&0.9229	&0.9386	&0.9373	&0.9319	&0.9334	&0.8026	&0.9503	&0.9483\\
	                     &GBDT	&0.9369	&0.9414	&0.9382	&0.9439	&0.9317	&0.9425	&0.9438	&0.9433	&0.9525	&0.9543\\\hline
\multirow{3}{*}{ecoli2}	&CART	&0.8632	&0.8568	&0.8596	&0.8817	&0.7666	&0.8663	&0.8692	&0.8889	&0.8811	&0.8914\\
	                     & SVM	&0.9446	&0.9555	&0.9439	&0.9563	&0.9553	&0.9512	&0.9444	&0.8779	&0.9611	&0.9531\\
	                     &GBDT	&0.9398	&0.9561	&0.9468	&0.9501	&0.9277	&0.9347	&0.9534	&0.9549	&0.9563	&0.9648\\\hline
\multirow{3}{*}{yeast3}	&CART	&0.8219	&0.8617	&0.8469	&0.8704	&0.8363	&0.8971	&0.8799	&0.8607	&0.9131	&0.8889\\
	                     &SVM	&0.9752	&0.9727	&0.9749	&0.9674	&0.9733	&0.9575	&0.9761	&0.7851	&0.9761	&0.9763\\
	                     &GBDT	&0.9615	&0.9638	&0.9596	&0.9685	&0.9544	&0.9612	&0.9625	&0.9666	&0.9682	&0.9687\\\hline
\multirow{3}{*}{page-blocks0}	&CART   	&0.9219	&0.9301	&0.9293	&0.9324	&0.9267	&0.9284	&0.9503	&0.9315	&0.9393	&0.9434\\
	                           &SVM	     &0.9079	&0.8961	&0.9089	&0.9204	&0.8631	&0.9132	&0.7696	&0.3164	&0.9273	&0.9189\\
	                           &GBDT     &0.9903	&0.9914	&0.9908	&0.9896	&0.9889	&0.9919	&0.9916	&0.9904	&0.9921	&0.9915\\\hline
\multirow{3}{*}{yeast-2\_vs\_4}	&CART	     &0.8388	&0.8695	&0.8299	&0.8216	&0.9065	&0.9105	&0.8892	&0.8374	&0.9247	&0.9452\\
	                           &SVM	     &0.9592	&0.9574	&0.9583	&0.9636	&0.9644	&0.9683	&0.9549	&0.7821	&0.9736	&0.9765\\
	                           &GBDT	&0.9821	&0.9805	&0.9812	&0.9474	&0.9713	&0.9768	&0.9768	&0.9738	&0.9814	&0.9825\\\hline
\multirow{3}{*}{abalone9-18}	&CART	     & 0.6422	&0.6917	&0.6271	&0.6311	&0.6905	&0.7303	&0.7118	&0.7127	&0.7442	&0.6947\\
	                           &SVM	     & 0.8987	&0.8931	&0.9007	&0.9189	&0.8377	&0.9241	&0.6759	&0.2664	&0.9105	&0.9297\\
	                           &GBDT	&0.8166	&0.8344	&0.8119	&0.8305	&0.8349	&0.8066	&0.7884	&0.8294	&0.8499	&0.8424\\\hline
\multirow{3}{*}{winequality-red-4}	&CART	     &0.5485	&0.5812	&0.5372	&0.5402	&0.5735	&0.6096	&0.6564	&0.5508	&0.6898	&0.6406\\
	                           &SVM	     &0.3242	&0.5919	&0.4069	&0.4845	&0.5364	&0.6059	&0.5705	&0.3997	&0.5914	&0.6317\\
	                           &GBDT	&0.7161	&0.6996	&0.7174	&0.7186	&0.6966	&0.7288	&0.7309	&0.7172	&0.7448	&0.7415\\\hline
\multirow{3}{*}{yeast5}	&CART	&0.8687	&0.9126	&0.8548	&0.9356	&0.8822	&0.9253	&0.8906	&0.8862	&0.9283	&0.9685\\
	                     & SVM	&0.9885	&0.9883	&0.9788	&0.9784	&0.7879	&0.9813	&0.9781	&0.8386	&0.9688	&0.9885\\
	                     &GBDT	&0.9686	&0.9856	&0.9866	&0.9833	&0.9668	&0.9816	&0.9791	&0.9754	&0.9856	&0.9869\\\hline
\multirow{3}{*}{winequality-red-8\_vs\_6}	&CART	     &0.6617	&0.6975	&0.6859	&0.6101	&0.7389	&0.7589	&0.7679	&0.7732	&0.7669	&0.7818\\
	&SVM	&0.5623	&0.6377	&0.5406	&0.5455	&0.5621	&0.7158	&0.5439	&0.3924	&0.6488	&0.6144\\
	&GBDT	&0.8918	&0.8279	&0.8997	&0.8911	&0.8949	&0.9066	&0.8269	&0.8336	&0.9158	&0.9047\\\hline
\multirow{3}{*}{abalone-17\_vs\_7-8-9-10}	&CART	     & 0.6036	&0.6372	&0.6271	&0.6731	&0.6001	&0.7224	&0.8747	&0.8456	&0.7593	&0.8304\\
	&SVM	&0.9051	&0.9401	&0.9085	&0.9417	&0.7022	&0.9442	&0.8005	&0.6894	&0.9459	&0.9467\\
	&GBDT	&0.8925	&0.9074	&0.8991	&0.9077	&0.7959	&0.9105	&0.8998	&0.8909	&0.9075	&0.9149\\\hline
\multirow{3}{*}{abalone-21\_vs\_8}	     &CART	     & 0.7947	&0.8474	&0.7938	&0.7438	&0.6088	&0.8239	&0.8262	&0.8297	&0.8598	&0.8758\\
	&SVM	&0.9152	&0.9129	&0.9152	&0.8877	&0.7805	&0.8997	&0.9156	&0.4028	&0.9156	&0.9207\\
	&GBDT	&0.9051	&0.9122	&0.9024	&0.8899	&0.6145	&0.9012	&0.8949	&0.8861	&0.9073	&0.9213\\\hline
\multirow{3}{*}{winequality-white-3\_vs\_7}	     &CART	     &0.6705	&0.5619	&0.6449	&0.6693	&0.6812	&0.7432	&0.7074	&0.6835	&0.6915	&0.8023\\
	&SVM	&0.7216	&0.8511	&0.7256	&0.7341	&0.8421	&0.8789	&0.7642	&0.5449	&0.8461	&0.8839\\
	&GBDT	&0.8185	&0.8324	&0.8293	&0.8173	&0.7926	&0.8698	&0.8776	&0.7642	&0.8213	&0.8872\\\hline
\multirow{3}{*}{abalone-20\_vs\_8-9-10}	&CART	     &0.5778	&0.6831	&0.6364	&0.6033	&0.5021	&0.7747	&0.8199	&0.6962	&0.8144	&0.7857\\
	&SVM	&0.9439	&0.9561	&0.9551	&0.9351	&0.7547	&0.9594	&0.7633	&0.2646	&0.9575	&0.9661\\
	&GBDT	&0.9081	&0.9071	&0.8811	&0.9008	&0.5249	&0.8931	&0.8955	&0.8862	&0.9116	&0.8988\\\hline
\multirow{3}{*}{poker-8-9\_vs\_5}	     &CART	     &0.5844	&0.5391	&0.6106	&0.6113	&0.5621	&0.7197	&0.6428	&0.7159	&0.7223	&0.6768\\
	&SVM	&0.8942	&0.8557	&0.8986	&0.8974	&0.3472	&0.8689	&0.5766	&0.3452	&0.8155	&0.8708\\
	&GBDT	&0.8361	&0.7681	&0.8332	&0.8721	&0.7507	&0.8731	&0.7128	&0.6763	&0.8459	&0.8874\\\hline
\multirow{3}{*}{abalone19}	&CART	&0.5458	&0.5488	&0.5631	&0.5403	&0.4923	&0.5675	&0.6902	&0.5831	&0.6921	&0.6926\\
	&SVM	&0.5422	&0.7398	&0.5909	&0.5159	&0.3779	&0.7081	&0.7173	&0.3108	&0.7433	&0.6772\\
	&GBDT	&0.7961	&0.7727	&0.7977	&0.7766	&0.4632	&0.7676	&0.7741	&0.7877	&0.8198	&0.8392\\\hline		 
	\end{tabular}}
\end{table}

\begin{table}[!htbp]
	\centering
	\label{tb6}
	\caption{suPR of all algorithms in the collected datasets}
	\resizebox{\textwidth}{110mm}{
	\begin{tabular}{cccccccccccc}
		\hline
dataset	&models	&Null	&SMOTE	&TL	&NB-TL	&NB-Comm	&UCBSS	&SPE	&RUSBoost	&MGRU-MD	&MGRU-SED \\ \hline
\multirow{3}{*}{pima}	&CART&0.5583	&0.5974	&0.6005	&0.7041	&0.6881	&0.7017	&0.6884	&0.6311	&0.6983	&0.7112\\
                     &SVM	&0.7781	&0.7779	&0.7771	&0.7758	&0.7701	&0.7786	&0.7767	&0.6161	&0.7779	&0.7874\\
                     &GBDT	&0.7774	&0.7782	&0.7787	&0.7778	&0.7751	&0.7794	&0.7782	&0.7795	&0.7769	&0.7877\\\hline
\multirow{3}{*}{glass0}	&CART	&0.6612	&0.6991	&0.7201	&0.7238	&0.7454	&0.7674	&0.7858	&0.7645	&0.7725	&0.7401\\
	                     &SVM	&0.6892	&0.6286	&0.7891	&0.7792	&0.7078	&0.7818	&0.7161	&0.6393	&0.7978	&0.7859\\
	                     &GBDT	&0.8065	&0.8062	&0.8079	&0.8029	&0.7791	&0.8077	&0.8056	&0.7933	&0.8168	&0.8051\\\hline
\multirow{3}{*}{haberman}	&CART	&0.5602	&0.5655	&0.5551	&0.6244	&0.6199	&0.6925	&0.7029	&0.5938	&0.6191	&0.6032\\
	                     &SVM	&0.7969	&0.8022	&0.8007	&0.7997	&0.7345	&0.7839	&0.7754	&0.6099	&0.7946	&0.8118\\
	                     &GBDT	&0.7856	&0.7827	&0.7882	&0.7716	&0.7836	&0.7854	&0.7968	&0.7977	&0.7893	&0.8076\\\hline
\multirow{3}{*}{vehicle1}	&CART	&0.6164	&0.6571	&0.6284	&0.7301	&0.6954	&0.7115	&0.7998	&0.7999	&0.7157	&0.7953\\
	                     &SVM	&0.8349	&0.8311	&0.8404	&0.8281	&0.8238	&0.8457	&0.8236	&0.6356	&0.8544	&0.8449\\
	                     &GBDT	&0.8485	&0.8489	&0.8486	&0.8456	&0.8441	&0.8508	&0.8482	&0.8476	&0.8485	&0.8543\\\hline
\multirow{3}{*}{vehicle0}	&CART	&0.8051	&0.8309	&0.8101	&0.8308	&0.8344	&0.8316	&0.8726	&0.8543	&0.8604	&0.8414\\
	                     &SVM	&0.8654	&0.8695	&0.8645	&0.8625	&0.8658	&0.8715	&0.8658	&0.6388	&0.8755	&0.8704\\
	                     &GBDT	&0.8715	&0.8717	&0.8715	&0.8716	&0.8668	&0.8717	&0.8716	&0.8715	&0.8716	&0.8733\\\hline
\multirow{3}{*}{ecoli1}	&CART	&0.7674	&0.7724	&0.7938	&0.8282	&0.8153	&0.7937	&0.8443	&0.8439	&0.8449	&0.8509\\
	                     &SVM	&0.8719	&0.8708	&0.8715	&0.8718	&0.8718	&0.8734	&0.8728	&0.8105	&0.8824	&0.8825\\
	                     &GBDT	&0.8726	&0.8728	&0.8724	&0.8728	&0.8714	&0.8737	&0.8719	&0.8709	&0.8831	&0.8836\\\hline
\multirow{3}{*}{ecoli2}	&CART	&0.8208	&0.8010	&0.8101	&0.8435	&0.8597	&0.8715	&0.8478	&0.8587	&0.8909	&0.8402\\
	                     &SVM	&0.9145	&0.9166	&0.9151	&0.9159	&0.9161	&0.9162	&0.9136	&0.8776	&0.9255	&0.9159\\
	                     &GBDT	&0.9151	&0.9167	&0.9159	&0.9168	&0.9161	&0.9173	&0.9168	&0.9168	&0.9261	&0.9162\\\hline
\multirow{3}{*}{yeast3}	&CART	&0.7862	&0.8266	&0.8136	&0.8393	&0.8536	&0.9043	&0.9001	&0.9097	&0.9134	&0.9008\\
	                     &SVM	&0.9411	&0.9423	&0.9424	&0.9425	&0.9423	&0.9424	&0.9424	&0.7648	&0.9436	&0.9424\\
	                     &GBDT	&0.9316	&0.9417	&0.9418	&0.9416	&0.9421	&0.9423	&0.9419	&0.9422	&0.9432	&0.9422\\\hline
\multirow{3}{*}{page-blocks0}	&CART	     &0.8791	&0.8935	&0.8871	&0.9036	&0.9025	&0.8889	&0.9154	&0.9102	&0.9098	&0.9224\\
	&SVM	&0.9433	&0.9427	&0.9432	&0.9443	&0.9353	&0.9432	&0.9351	&0.6291	&0.9417	&0.9468\\
	&GBDT	&0.9469	&0.9471	&0.9461	&0.9460	&0.9469	&0.9469	&0.9469	&0.9469	&0.9451	&0.9482\\\hline
\multirow{3}{*}{yeast-2\_vs\_4}	&CART	     &0.8039	&0.8424	&0.7938	&0.7889	&0.8894	&0.9419	&0.9287	&0.9391	&0.9433	&0.9198\\
	&SVM	&0.9469	&0.9483	&0.9481	&0.9477	&0.9481	&0.9479	&0.9475	&0.8257	&0.9498	&0.9479\\
	&GBDT	&0.9482	&0.9483	&0.9372	&0.9472	&0.9481	&0.9425	&0.9481	&0.9483	&0.9484	&0.9483\\\hline
\multirow{3}{*}{abalone9-18}	&CART	     &0.6439	&0.7176	&0.6282	&0.6306	&0.7359	&0.8011	&0.9436	&0.8659	&0.8786	&0.7778\\
	&SVM	&0.9686	&0.9685	&0.9688	&0.9692	&0.9666	&0.9693	&0.9563	&0.7508	&0.9702	&0.9693\\
	&GBDT	&0.9663	&0.9674	&0.9661	&0.9659	&0.9665	&0.9655	&0.9555	&0.9568	&0.9689	&0.9676\\\hline
\multirow{3}{*}{winequality-red-4}&CART	&0.5608	&0.6107	&0.5579	&0.5586	&0.6443	&0.6954	&0.8803	&0.7684	&0.8029	&0.7429\\
	&SVM	&0.9281	&0.9149	&0.9457	&0.9487	&0.9533	&0.9526	&0.9447	&0.9071	&0.9461	&0.9541\\
	&GBDT	&0.9764	&0.9748	&0.9725	&0.9746	&0.9742	&0.9774	&0.9761	&0.9749	&0.9765	&0.9804\\\hline
\multirow{3}{*}{yeast5}&CART	&0.8576	&0.9026	&0.8451	&0.9276	&0.8925	&0.9351	&0.9103	&0.9098	&0.9301	&0.9726\\
	&SVM	&0.9849	&0.9849	&0.9849	&0.9849	&0.9231	&0.9850	&0.9849	&0.7536	&0.9851	&0.9851\\
	&GBDT	&0.9601	&0.9849	&0.9851	&0.9849	&0.9847	&0.9850	&0.9849	&0.9726	&0.9851	&0.9849\\\hline
\multirow{3}{*}{winequality-red-8\_vs\_6}	&CART	     &0.6751	&0.7251	&0.6939	&0.6281	&0.8158	&0.8642	&0.9137	&0.7219	&0.9159	&0.8645\\
	&SVM	&0.9701	&0.9772	&0.9629	&0.9776	&0.8768	&0.9691	&0.9539	&0.8978	&0.9695	&0.9306\\
	&GBDT	&0.9821	&0.9731	&0.9851	&0.9847	&0.9845	&0.9854	&0.9839	&0.9842	&0.9855	&0.9851\\\hline
\multirow{3}{*}{abalone-17\_vs\_7-8-9-10}	&CART	     &0.6016	&0.6504	&0.6262	&0.6759	&0.7035	&0.8204	&0.9118	&0.9019	&0.8053	&0.9187\\
	&SVM	&0.9868	&0.9872	&0.9868	&0.9871	&0.9321	&0.9871	&0.9852	&0.8533	&0.9871	&0.9872\\
	&GBDT	&0.9861	&0.9871	&0.9861	&0.9868	&0.9836	&0.9868	&0.9867	&0.9866	&0.9868	&0.9769\\\hline
\multirow{3}{*}{abalone-21\_vs\_8}&CART   &0.7649	&0.9164	&0.7948	&0.7465	&0.8189	&0.9079	&0.9156	&0.8914	&0.9126	&0.9164\\
	&SVM	&0.9848	&0.9864	&0.9853	&0.9838	&0.9813	&0.9855	&0.9861	&0.8739	&0.9868	&0.9849\\
	&GBDT	&0.9842	&0.9866	&0.9841	&0.9603	&0.9788	&0.9867	&0.9852	&0.9325	&0.9861	&0.9869\\\hline
\multirow{3}{*}{winequality-white-3\_vs\_7}&CART	&0.6705	&0.5797	&0.6501	&0.6727	&0.7406	&0.7635	&0.8571	&0.8456	&0.7405	&0.8635\\
	&SVM	&0.9179	&0.9863	&0.9539	&0.9728	&0.9869	&0.9677	&0.9388	&0.7284	&0.9871	&0.9874\\
	&GBDT	&0.8408	&0.9864	&0.8632	&0.8657	&0.9861	&0.9817	&0.9827	&0.9856	&0.9868	&0.9876\\\hline
\multirow{3}{*}{abalone-20\_vs\_8-9-10}	&CART	     &0.5855	&0.6957	&0.6417	&0.6089	&0.6823	&0.8513	&0.8945	&0.7051	&0.8844	&0.9095\\
	&SVM	&0.9928	&0.9931	&0.9929	&0.9929	&0.9449	&0.9931	&0.9909	&0.9435	&0.9930	&0.9931\\
	&GBDT	&0.9922	&0.9913	&0.9912	&0.9924	&0.9871	&0.9925	&0.9926	&0.9925	&0.9923	&0.9927\\\hline
\multirow{3}{*}{poker-8-9\_vs\_5}&CART	&0.5924	&0.5606	&0.6164	&0.6174	&0.6353	&0.8152	&0.9172	&0.9339	&0.8281	&0.7905\\
	&SVM	&0.9909	&0.9431	&0.9927	&0.9931	&0.9047	&0.9929	&0.9808	&0.8971	&0.9909	&0.9932\\
	&GBDT	&0.9928	&0.9893	&0.9928	&0.9931	&0.9920	&0.9928	&0.9853	&0.9889	&0.9922	&0.9931\\\hline
\multirow{3}{*}{abalone19}&CART	&0.5517	&0.5684	&0.5676	&0.5473	&0.6275	&0.7759	&0.8956	&0.8996	&0.7957	&0.7999\\
	&SVM	&0.9892	&0.9956	&0.9927	&0.9898	&0.9868	&0.9942	&0.9948	&0.9434	&0.9958	&0.9931\\
	&GBDT	&0.9945	&0.9952	&0.9953	&0.9953	&0.9879	&0.9947	&0.9951	&0.9951	&0.9955	&0.9958\\\hline		 
	\end{tabular}}
\end{table}

\begin{figure}[!htbp]
	\centering
	\includegraphics[scale=1]{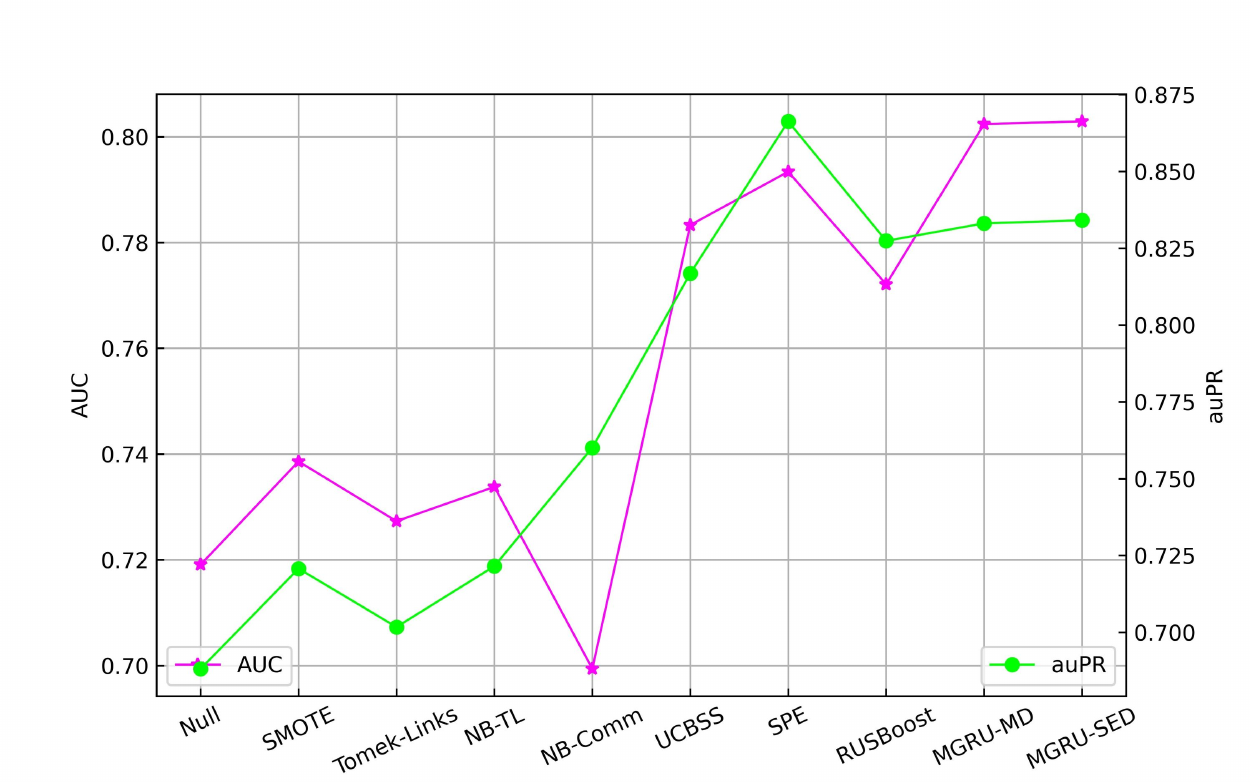}
	\caption{The mean value of CART on all experimental datasets.}
	\label{fig3}
\end{figure}

\begin{figure}[!htbp]
	\centering
	\includegraphics[scale=1]{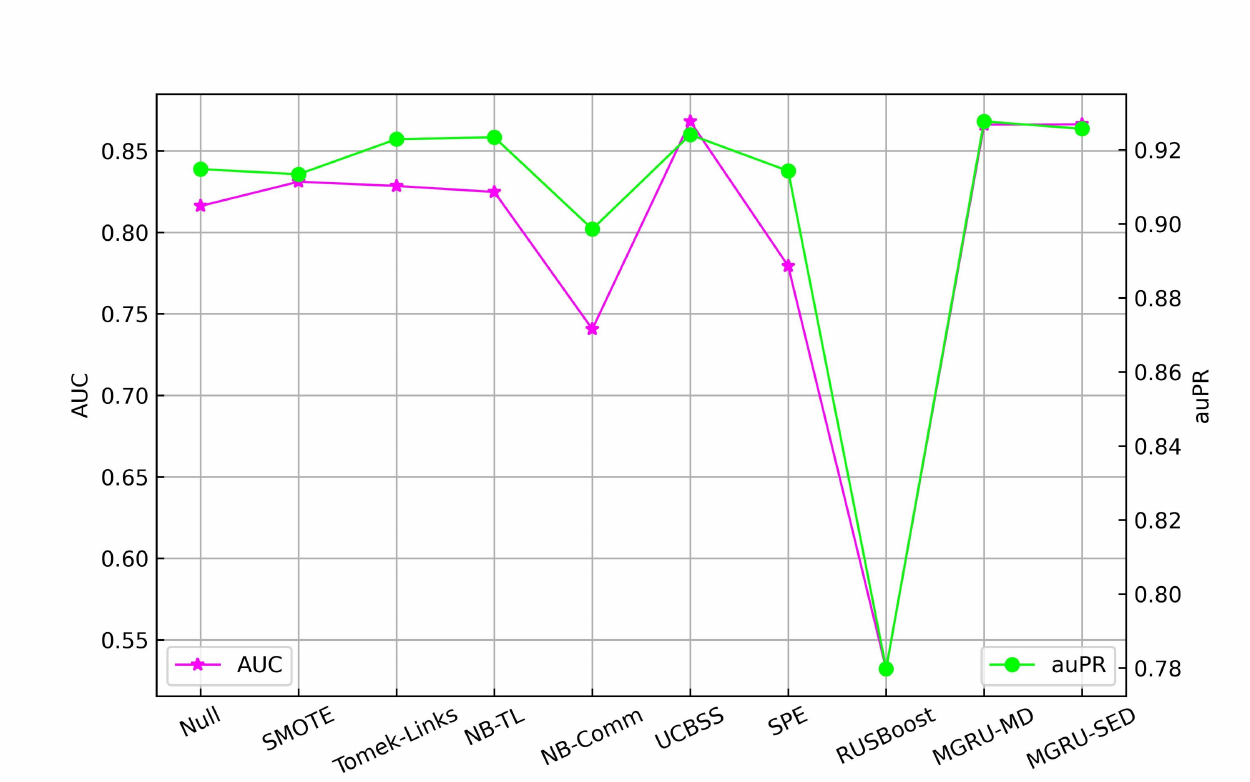}
	\caption{The mean value of SVM on all experimental datasets.}
	\label{fig4}
\end{figure}

\begin{figure}[!htbp]
	\centering
	\includegraphics[scale=1]{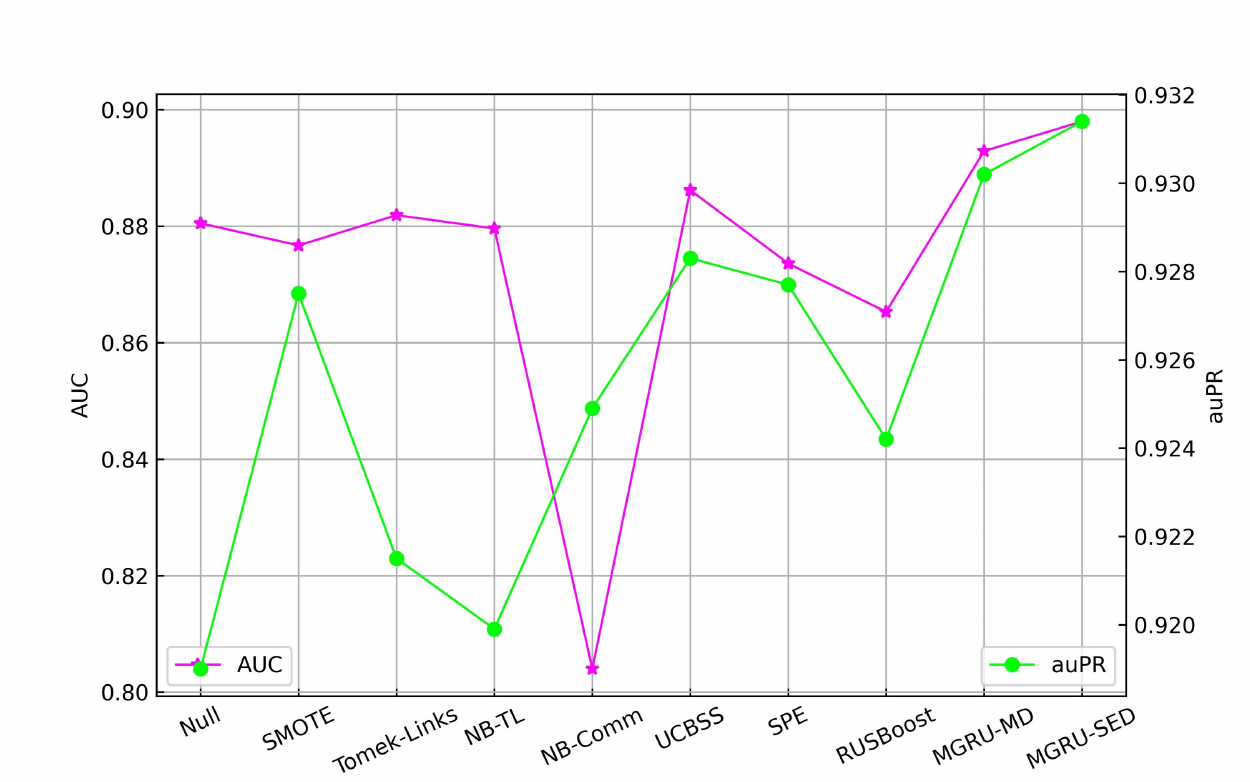}
	\caption{The mean value of GBDT on all experimental datasets.}
	\label{fig5}
\end{figure}

\begin{table}[!htbp]
	\centering
	\caption{Average Friedman rankings of and APVs using Holm’s procedure in AUC.}
	\begin{tabular}{cccc}
		\hline
		Result (AUC)	&Algorithm	&Friedman ranking	&APVs \\ \hline
		\multirow{10}{*}{CART} &MGRU-MD	&2.500	&-\\
&MGRU-SED	&2.875	&0.695299\\
&SPE	&3.400	&0.694415\\
&UCBSS	&3.975	&0.370252\\
&RUSBoost	&4.450	&0.166717\\
&NB-TL	&6.550	&0.000117\\
&SMOTE	&6.650	&0.000088\\
&Tomek-Links	&7.925	&0\\
&NB-Comm	&8.275	&0\\
&Null	&8.400	&0\\\hline
\multirow{10}{*}{SVM} &MGRU-SED	&2.325	&-\\
&MGRU-MD	&2.650	&0.734270\\
&UCBSS	&3.700	&0.301925\\
&SMOTE	&5.100	&0.011252\\
&NB-TL	&5.650	&0.002060\\
&Tomek-Links	&5.875	&0.001131\\
&Null	&5.900	&0.001131\\
&SPE	&6.700	&0.000034\\
&NB-Comm	&7.200	&0.000003\\
&RUSBoost	&9.900	&0\\\hline
\multirow{10}{*}{GBDT} &MGRU-SED	&2.275	&-\\
&MGRU-MD	&2.650	&0.695299\\
&UCBSS	&4.925	&0.011286\\
&SMOTE	&5.150	&0.008024\\
&Tomek-Links	&5.650	&0.001736\\
&SPE	&5.700	&0.001736\\
&Null	&6.150	&0.000311\\
&NB-TL	&6.250	&0.000231\\
&RUSBoost	&7.250	&0.000002\\
&NB-Comm	&9.000	&0\\\hline
	\end{tabular}
\end{table}

\begin{table}[!htbp]
	\centering
	\caption{Average Friedman rankings of and APVs using Holm’s procedure in auPR.}
	\begin{tabular}{cccc}
		\hline
		Result (auPR)	&Algorithm	&Friedman ranking	&APVs \\ \hline
		\multirow{10}{*}{CART} &SPE	&2.550	&-\\
&MGRU-MD	&3.000	&1\\
&MGRU-SED	&3.175	&1\\
&RUSBoost	&3.800	&0.575084\\
&UCBSS	&4.200	&0.339287\\
&NB-Comm	&5.800	&0.003438\\
&NB-TL	&6.950	&0.000026\\
&SMOTE	&7.575	&0.000001\\
&Tomek-Links	&8.750	&0\\
&Null	&9.200	&0\\\hline
\multirow{10}{*}{SVM} &MGRU-MD	&2.725	&-\\
&MGRU-SED	&2.900	&0.854969\\
&UCBSS	&3.700	&0.617019\\
&SMOTE	&4.825	&0.084840\\
&NB-TL	&5.125	&0.048743\\
&Tomek-Links	&5.375	&0.028215\\
&Null	&6.400	&0.000743\\
&SPE	&6.800	&0.000146\\
&NB-Comm	&7.250	&0.000018\\
&RUSBoost	&9.900	&0\\\hline
\multirow{10}{*}{GBDT} &MGRU-SED	&2.300	&-\\
&MGRU-MD	&3.500	&0.210075\\
&UCBSS	&4.100	&0.120206\\
&SMOTE	&5.100	&0.010350\\
&SPE	&6.000	&0.000445\\
&Tomek-Links	&6.300	&0.000157\\
&RUSBoost	&6.325	&0.000157\\
&NB-TL	&6.500	&0.000081\\
&Null	&7.000	&0.000007\\
&NB-Comm	&7.875	&0\\\hline
	\end{tabular}
\end{table}

Throughout the experiment, we will focus on the performance differences between the proposed MGRU-MD and MGRU-SED algorithms and other baseline algorithms. According to Table 5 and Table 6, we can make a simple statistics. When we choose CART as the base classifier, MGRU-MD and MGRU-SED obtain the best AUC values on 6 data sets and 8 data sets, respectively. When we observe the auPR evaluation metric, the number of data sets that MGRU-SED obtains the optimal value is significantly more than that of MGRU-MD. However, the number of optimal values obtained by other classification algorithms on all data sets is very small. When we continue to observe the performance difference between the ensemble learning and the proposed two algorithms, we can conclude that the baseline ensemble learning algorithm SPE is our main competitor. We simply count the number of optimal values obtained by each algorithm on all data sets. SPE obtains the optimal AUC value on 6 data sets, and the optimal auPR value on 5 data sets. From the point of view of the optimal number of values obtained from all data sets, the performance of the proposed MGRU-MD and MGRU-SED is equivalent to that of SPE. RUSBoost is an ensemble learning algorithm based on random undersampling, which has not obtained the optimal value on all data sets. Therefore, in terms of obtaining the optimal number of values, the classification performance of RUSBoost is not as good as our proposed algorithm.

Next, we can perform a brief analysis based on the mean results of each algorithm on all data sets in Fig.3. For the AUC value, our proposed MGRU-SED is the best, followed by MGRU-MD, and followed by SPE. However, under the auPR evaluation metric, the two algorithms we proposed have a significant gap compared with SPE. The main reason is that the two algorithms we proposed remove partially overlapping majority instances before training, which may lead to a decrease in the accuracy of majority instances, resulting in a decrease in the overall auPR value of the model. On the whole, we can draw a simple conclusion that our proposed algorithm is superior to other resampling algorithms on most data sets. Compared with other ensemble learning algorithms, our algorithm has a competitive advantage.

When we select SVM as the base classifier, the number of optimal values obtained on all data sets in Table 5 and Table 6 or the mean value of algorithms on all data sets are depicted in Fig. 4, we can conclude that the proposed MGRU-SED and MGRU-MD have better performance than the state-of-the-art class-overlap under-sampling algorithm. SPE does not directly balance all training data sets, but eliminates majority instances one by one through the use of self-paced learning. Therefore, the SVM may not be able to obtain the optimal performance every time it is learned. When the performance of the obtained SVM is poor, it will affect the final integrated performance of the model. Many researchers believe that RUS may delete a large number of valuable majority samples. In the random process, if valuable majority instances are deleted, the classification performance of the model will be greatly reduced. Although RUSBoost, an ensemble learning algorithm based on RUS, can effectively improve the classification performance of the model, it still cannot overcome this shortcoming. Especially for SVM, if majority instances we delete are support vectors, the performance will drop off a cliff. Compared with other resampling algorithms, although UCBSS has the highest mean value under the AUC evaluation metric, the two algorithms we proposed follow closely behind, and the gap is very small. Therefore, we can conclude that when we use SVM for classification, under the two evaluation metrics of AUC and auPR, the performance of MGRU-SED and MGRU-MD is better than or partly better than other resampling algorithms.

When we choose GBDT as the base classifier, according to the results in Table 5 and Table 6, we can observe that the two algorithms proposed by us can obtain excellent classification performance compared with other algorithms, whether in metrics of AUC or auPR. In particular, MGRU-SED has better performance.

In order to verify the above analysis results, Tables 7 and 8 show the average Friedman rankings and APVs values of all comparison algorithms under the two metrics of AUC and auPR. We can observe that the two algorithms MGRU-SED and MGRU-MD can get the lowest value in most cases. When we used the CART classifier, SPE obtained the lowest value under the evaluation index of auPR. Although our proposed algorithm does not occupy an absolute advantage under the above circumstances, we can still draw the conclusion that the two proposed algorithms are excellent in most cases according to the Friedman ranking results. 

Our further analysis of the results of Holm’s post-hoc test shows that for AUC, when CART is selected for classification, the APVs values of the five algorithms of MGRU-MD, MGRU-SED, SPE, UCBSS and RUSBoost are all greater than the commonly used significance level (e.g. $\alpha=0.05$). Therefore, we believe that in this case, there is no significant difference in the performance of these five algorithms. When selecting SVM for classification, the APVs of the three algorithms of MGRU-SED, MGRU-MD and UCBSS are greater than 0.05, so there is no significant difference in performance between them. When we choose GBDT for classification, the APVs values corresponding to all algorithms are lower than the significance level $\alpha=0.05$. Therefore, we believe that in this case, the null hypothesis of equality is rejected, which also supports the conclusion of our proposed algorithm is superior to other algorithms. For auPR, our conclusions are the same as those under the AUC.

In summary, based on the experimental results and statistical test results, we know that the two algorithms we proposed are superior to other classification algorithms in most cases. Similarly, the Friedman ranking and holm’s post-hoc test results obtained in Tables 7 and 8 also support our conclusions.

\subsection{Global analysis of results}

Finally, we can make a global analysis of results combining the results offered by Tables from 3-7 and Figure from 2-5:

(1)When we use $ONB_{avg}$ to calculate the complexity of the training set after sampling, according to the results in Table 3 and Fig. 2, we can conclude that the proposed MGRU-SED and MGRU-MD are the best algorithms, and MGRU-SED is better. Therefore, we believe that the two proposed algorithms can effectively reduce the class overlap complexity of the training set. Their performance is better than other algorithms, and this hypothesis has been confirmed by non-parametric statistical tests.

(2)$ONB_{avg}$ is a conservative measure of class-overlap complexity. When we measure the complexity of oversampling algorithms such as SMOTE, we cannot accurately explain the exact overlap complexity of such algorithms, and future work can still improve it. In addition, as described in \cite{61pascual2021revisiting}, the process of searching the hyperspheres in $ONB_{avg}$ is NP-hard. For moderately large datasets, the calculation time is prohibitively long.

(3)According to the experimental results in Table 5-6, the two algorithms we proposed are superior to other state-of-the-arts class-overlap undersampling algorithms in most cases. This conclusion was confirmed by non-parametric statistical tests.

(4)When we choose CART as the base classifier, according to the experimental results in Fig. 3, SPE's auPR is superior to all class-overlap undersampling algorithms including our proposed algorithm, and its performance is improved thanks to the use of an integrated learning framework. Of course, our algorithm can also further improve the classification performance utilizing the MGRU class-overlap under-sampling algorithms through ensemble learning.

(5)According to the results in Table 5-6 and Fig. 4, when we choose SVM as the base classifier, the classification performance is better on a more balanced data set. In addition, for a highly imbalanced dataset, RUSBoost deletes too many majority instances, which results in SVM not being able to better obtain the knowledge in the training data set, so performance is significantly reduced.

(6)For GBDT, our proposed algorithm is significantly better than other state-of-the-arts class-overlap undersampling algorithms. It is also better than the two ensemble learning methods of SPE and RUSBoost. This conclusion is confirmed by the non-parametric statistical tests in Table 7-8. Since our proposed algorithm has the ability to mine potential overlapping instances, it can eliminate the potential overlapping majority instances that are difficult to find by other algorithms. Therefore, in the learning process of GBDT, a classification model with excellent performance and robustness can be obtained.

(7)From the overall results, the performance of our proposed algorithm is better and can effectively improve the performance of the model. In some cases, TL, NB-TL and NB-Comm did not significantly improve the overall classification performance of the model. On the contrary, it will cause the overall performance of the classification model to decrease because of ignoring potential overlapping samples.

\section{Conclusions and future works}

The imbalanced data sets widely exist in various real-world data sets. For the problem of imbalanced data classification, some data-level algorithms have been proposed to solve such problems. In this paper, we consider the potential overlapping instances in the data set through the local subspace, and propose two novel MGRU-MD and MGRU-SED class-overlap under-sampling algorithms. The experimental results on 20 highly imbalanced datasets show that the AUC and auPR of the two class-overlap under-sampling algorithms we proposed are better or partially better than other state-of-the-arts resampling algorithms. Using the MGRU algorithm to preprocess the training set can significantly improve the classification performance of the model. In addition, according to non-parametric statistical tests, the performance difference between the two under-sampling methods, MGRU-SED and MGRU-MD, is not significant. When the number of instances in the data set is less than the number of features, we will not be able to calculate the Mahalanobis distance. Therefore, we can preferentially choose MGRU-SED for under-sampling.

In the design process of the MGRU algorithm, we did not consider whether different feature combinations may dig out more potential overlapping instances. Therefore, in the future work we will further explore whether there are potentially overlap instances in the local subspaces of different feature combinations. Compared with the SPE ensemble learning algorithm, we can find that removing majority instances will lead to a decrease in the accuracy of majority instances. Therefore, we can further combine the ensemble learning framework to propose an ensemble learning model under multi-granularity local subspaces to prevent the deletion of class-overlap instances from reducing the classification accuracy of majority instances. On the basis of our proposed algorithm, we can also explore the class-overlap undersampling algorithms of multi-class imbalanced data. Due to the essential difference between the multi-class classification problem and the binary classification problem, in future work, we can further apply this algorithm to the multi-class overlap problem to explore the classification performance of our proposed method in multi-class classification problem.

\section*{Acknowledgements}

This work was supported by the Science Foundation of China University of Petroleum, Beijing (No.2462020YXZZ023).

\bibliography{bibfile}

\end{document}